\documentclass[lettersize,journal]{IEEEtran}
\usepackage{amsmath,amsfonts}
\usepackage{array}
\usepackage{stfloats}
\usepackage{graphicx}
\usepackage{cite}
\usepackage[T1]{fontenc}
\usepackage{booktabs}
\usepackage{threeparttable}
\usepackage{microtype}
\usepackage{tikz}
\usetikzlibrary{arrows.meta,fit,calc,backgrounds}
\usepackage[hidelinks,bookmarksnumbered,bookmarksopen,bookmarksopenlevel=2,
            bookmarksdepth=2]{hyperref}
\graphicspath{{figures/}}

\newcommand{\method}{SCDF}
\newcommand{\supp}[1]{Supplement~S\nobreakdash-#1}
\newcommand{\px}{\,px}
\newcommand{\excluded}{$^\dagger$}

\definecolor{scdfink}{HTML}{1A1A1A}      % body ink
\definecolor{scdfblue}{HTML}{0072B2}     % regular pipeline stage
\definecolor{scdfgreen}{HTML}{009E73}    % inputs / outputs
\definecolor{scdfselfcal}{HTML}{D55E00}  % SELF-CALIBRATING (the core claim)
\definecolor{scdfgrey}{HTML}{8C8C8C}     % carried state
\definecolor{scdfstagefill}{HTML}{E7F0F7}
\definecolor{scdfsteprule}{HTML}{9DB8CC}

\tikzset{
  scdf/.style={
    font=\sffamily\scriptsize,
    text=scdfink,
    line width=0.7pt,
    inner sep=3pt,
    node distance=3mm,
  },
  phasebox/.style={
    draw=scdfblue, fill=scdfstagefill, rounded corners=4pt,
    line width=0.9pt, inner sep=6pt,
  },
  phasetitle/.style={
    font=\sffamily\small\bfseries, text=scdfblue, align=left,
  },
  panel/.style={
    draw=scdfsteprule, fill=white, rounded corners=3pt,
    line width=0.7pt, inner sep=0pt,
  },
  paneltitle/.style={
    font=\sffamily\scriptsize\bfseries, text=scdfink, align=left,
  },
  panelcap/.style={font=\sffamily\tiny, text=scdfink!75, align=center},
  stepnum/.style={
    circle, fill=scdfblue, text=white, font=\sffamily\tiny\bfseries,
    inner sep=0pt, minimum size=3.1mm,
  },
  gridline/.style={draw=scdfgreen!58!black, line width=0.4pt},
  cellfill/.style={draw=scdfgreen!62!black, fill=scdfgreen!35, line width=0.5pt},
  winbox/.style={draw=scdfselfcal, line width=0.7pt,
    dash pattern=on 1.5pt off 1.1pt},
  minilbl/.style={font=\sffamily\tiny, text=scdfink!72, inner sep=1pt},
  flowarr/.style={-{Stealth[length=5pt,width=4pt]}, line width=1.1pt,
    draw=scdfink},
  thinarr/.style={-{Stealth[length=3pt,width=2.4pt]}, line width=0.5pt,
    draw=scdfink!70},
  looparr/.style={-{Stealth[length=4pt,width=3pt]}, line width=0.7pt,
    draw=scdfblue!70!black, dashed},
}

\DeclareRobustCommand{\stepdot}[1]{%
  \tikz[baseline=-0.62ex]{\node[circle, fill=scdfblue, text=white,
    font=\sffamily\tiny\bfseries, inner sep=0pt, minimum size=2.7mm] {#1};}}

\hypersetup{
  pdftitle={Self-Calibrating Dense Displacement Fields for Reliable
            Co-Registration of Large Optical Satellite Imagery},
  pdfauthor={Shoukun Sun and Zhe Wang and Sanaz Salati and Jiyin Zhang and
            Hui Wang and Xiaogang Ma},
}

\begin{document}

\title{Self-Calibrating Dense Displacement Fields for Reliable
       Co-Registration of Large Optical Satellite Imagery}

\author{Shoukun~Sun,
        Zhe~Wang,
        Sanaz~Salati,
        Jiyin~Zhang,
        Hui~Wang,
        and~Xiaogang~Ma% <-this % stops a space
\thanks{Shoukun Sun, Jiyin Zhang, and Xiaogang Ma are with the Department of
Computer Science, University of Idaho, Moscow, ID 83844 USA (e-mail:
ssun@uidaho.edu; jiyinz@uidaho.edu; max@uidaho.edu).}% <-this % stops a space
\thanks{Zhe Wang is with the National Center for Ecological Analysis and
Synthesis (NCEAS), University of California, Santa Barbara, CA 93101 USA
(e-mail: zwang@nceas.ucsb.edu).}% <-this % stops a space
\thanks{Sanaz Salati is with Research Computing and Data Services (RCDS),
University of Idaho, Moscow, ID 83844 USA (e-mail: ssalati@idaho.edu).}% <-this % stops a space
\thanks{Hui Wang is with the Department of Geography and Planning,
Appalachian State University, Boone, NC 28608 USA
(e-mail: wangh4@appstate.edu).}% <-this % stops a space
\thanks{\emph{(Corresponding author: Xiaogang Ma.)}}}

% Running heads.
\markboth{}%
{Sun \MakeLowercase{\textit{et al.}}: Self-Calibrating Dense Displacement
Fields for Reliable Co-Registration of Large Optical Satellite Imagery}

\maketitle

\begin{abstract}
Co-registration underlies nearly every multi-temporal and multi-sensor use
of optical satellite imagery, and operational products still carry
documented offsets well above the fraction-of-a-pixel scale at which
change detection, time series, and data fusion degrade. Real image pairs
differ along several axes at once (sensor response, scene content,
viewing geometry, resolution, mosaic seams), and the last of these is
not a single global motion. Existing tools embed a motion model and
constants tuned to their development data; a pair that fits is registered
precisely, while one that does not either fails to match or returns a
result wrong by tens of pixels with no failure reported. Learned
matchers add a GPU requirement and carry no accuracy guarantee outside
their training distribution. We present \method{} (self-calibrating
displacement fields), a training-free, GPU-free estimator whose motion
model is the dense per-pixel displacement field itself, so no scene
motion falls outside the model. A single predict--measure--filter loop
runs over a resolution pyramid: the accumulated field predicts where each
patch of the moving image falls in the reference, RootSIFT matching and a
correlation pass measure the displacement there to sub-pixel precision,
and filters whose thresholds are all calibrated on the image pair itself
decide what survives. One configuration, with no per-dataset
tuning, processes full $8192^2$ scenes on a single CPU core. On 584
constructed-ground-truth pairs built from real Sentinel-2, Landsat-8/9,
and NAIP imagery, against seven classical baselines and two zero-shot
pretrained matchers, \method{} registers every pair with zero failures,
reduces the best baseline's real-pair median end-point error from 6.83 to
4.17\,m, and cuts its 90th percentile from 17.8 to 7.77\,m.
\end{abstract}


% !TEX root = ../../SCDF.tex
\section{Introduction}
\label{sec:intro}

\IEEEPARstart{O}{ptical} satellite images of the same area, acquired at
different times or
by different sensors, do not land on a common ground grid: geolocation
error, terrain correction, and inconsistent ground control each leave a
geometric residual. Co-registration is the task of removing this residual
by estimating the correction that brings a \emph{moving} image into
agreement with a \emph{reference}, so that content at the same pixel
position lies at the same point on the ground. Performed to sub-pixel
accuracy, it is a prerequisite for most multi-temporal and
multi-sensor analysis: change detection, time series of surface
reflectance, data fusion, and mosaicking all degrade when corresponding
pixels are misaligned by even a fraction of a pixel. Real products carry
documented misregistration at this scale and above.
Landsat-8 and Sentinel-2 imagery were temporarily misaligned by up to 38\,m
due to inconsistent ground control \cite{storey2016note}; Sentinel-2 pairs
from different relative orbits have exhibited misregistration up to
2.8 pixels \cite{skakun2017coregistration}; and the Harmonized
Landsat--Sentinel product reports residuals of 6--19\,m even after
dedicated co-registration \cite{claverie2018hls}. Correcting such
offsets automatically, at full scene size, for arbitrary same-optical
pairs is a routine need, and in a specific sense it remains unsolved.

% Figure: page-1 teaser.
\begin{figure}[t]
  \centering
  \includegraphics[width=\linewidth]{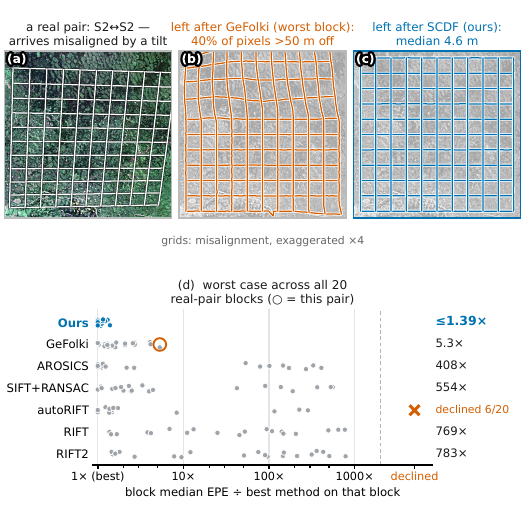}
  \caption{Misalignment drawn as a warped grid, exaggerated $4\times$.
  (a)~A real Sentinel-2 pair arrives misaligned by a constructed viewing
  tilt. (b)~What GeFolki, the baseline with the lowest real-pair median
  error, leaves behind at its worst benchmark block. (c)~What \method{}
  leaves on the same pair. (d)~Block-median error on the 20 real-pair
  blocks as a ratio to the best method per block ($\circ$ this block,
  $\times$ declined, right column the worst case); \method{} stays within
  $1.39\times$ everywhere.}
  \label{fig:teaser}
\end{figure}

The practical failure mode of existing tools is not imprecision but
\emph{conditional validity}. Each tool handles certain types of image
difference and fails severely on the rest. A real pair differs along
several axes at once: sensor radiometry and point-spread function,
scene content between dates, viewing geometry, ground sample distance,
and mosaicking seams, which correspond to no single global motion. The
conditional behavior follows from the motion model each tool family
embeds. Sparse feature matching fits a single global affine or homography;
phase-correlation grid tools assume translation within each window;
optical-flow and chip-tracking methods assume a small, smooth flow within
a fixed search range (Section~\ref{sec:related}). Each model covers some
axes and not others. On pairs its model covers, a tool is precise, and we
show that essentially every method is sub-pixel there. On a difference
type outside its model, it fails in one of two ways: it finds no
acceptable match and returns nothing, or it returns a field that is wrong
by tens of pixels, with nothing in the output flagging the failure. In
our benchmark, the baseline that wins the
most fixtures completes 554 of 584 problems, yet on 8 of its 20 real-pair
blocks the field it returns is $10\times$ or more from the best result
achieved on the same cells (Section~\ref{sec:results-bounded}). The
quantity that separates methods in practice is therefore not the precision
they reach when their model fits, but how far wrong they go on the
differences it does not cover. We call a method reliable when that worst
case is bounded, and we argue that the worst case across conditions
should replace peak accuracy as the primary reported axis.

A second, related problem is that standard accuracy reporting does not reveal it.
Per-pixel error distributions of failed or partially failed registrations are
strongly bimodal: a method can register half a scene perfectly and be
wrong by 100 pixels on the other half. The mean end-point error is
dominated by outliers to the point of being unusable, while the median
can remain sub-pixel for a field in which 44\% of pixels are grossly
wrong (Section~\ref{sec:benchmark-metrics}). Outlier-fraction metrics are long
standard in optical flow and matching benchmarks
(Section~\ref{sec:related}), but a fixed-threshold fraction cannot simply be
transplanted: pixel sizes in our benchmark range from 0.6 to 90\,m, so a
one-pixel tolerance is 0.6\,m of error on some pairs and 90\,m on others.
Our protocol instead scores each cell by order statistics of per-pixel error
in meters, keeps a declined cell in the aggregate as $+\infty$, and compares
methods by their distance from the best result achieved on each fixture.

Learned matchers do not resolve the reliability problem. Pretrained
detector and transformer matchers lead pairwise-matching benchmarks
\cite{sarlin2020superglue,lindenberger2023lightglue,sun2021loftr,edstedt2024roma},
but a trained model comes with no accuracy guarantee outside its training
distribution. This is conditional validity again, with the condition
implicit in the training set rather than explicit in a motion model. Extending
coverage to a new sensor pairing or scale ratio means pretraining or
fine-tuning on representative imagery. There is a separate operational cost: the dense and semi-dense models
run at small fixed working resolutions, so an $8192^2$ scene requires
tiled inference, a GPU, and a multi-gigabyte environment. We benchmark two of them zero-shot
(Section~\ref{sec:results}).

\textbf{Our answer}: We present \method{} (self-calibrating displacement
fields), a training-free dense co-registration method. It produces
candidate correspondences recall-first, then filters them with
measurement-quality gates and model-consistency tests whose thresholds
are all calibrated on the image pair itself. The output is a dense
per-pixel displacement field; no global motion model is imposed. The
pipeline has no absolute acceptance threshold anywhere: match acceptance is
gated against per-window \emph{null} correlations, error gates and outlier
tests are calibrated on the pair's own residual distributions, and the
final interpolator is selected by cross-validation on the pair's own
control points. One default configuration
is used for all 584 benchmark cells, with no per-dataset tuning. The
design predicts a results profile with no condition under which the
method degrades sharply (Section~\ref{sec:method-principle}).

\textbf{Contributions:}
\begin{enumerate}
  \item \textbf{Method (C1).} A three-phase dense displacement-field
  estimator (coarse-to-fine patch matching; phase-correlation sub-pixel
  re-measurement and lattice densification behind self-calibrating
  acceptance gates; cross-validated interpolation) that is training-free,
  matcher-agnostic, and runs arbitrary-size scenes.
  \item \textbf{Evaluation methodology (C2).} A benchmark of 584 cells with
  dense, verified-exact ground truth constructed on real multi-temporal and
  cross-sensor pairs (6 leg groups $\times$ 4 distortion families $\times$
  21--25 sites), and a scoring protocol built for it: errors in meters,
  declined cells kept in the aggregate as $+\infty$, per-cell tail quantiles
  in place of any pooled mean, and a ratio-to-best statistic that measures
  distance from what was achievable on each fixture
  (Section~\ref{sec:benchmark}).
  \item \textbf{Findings (C3).} The worst case separates methods;
  precision does not. With exact ground truth, every method whose model
  covers a fixture reaches sub-pixel accuracy on it. On real pairs,
  however, every baseline has at least one block where its worst-case error is
  5.3--783$\times$ the best result achieved there, or where it declines
  cells; \method{} produces a field on all 584 cells, ranks
  top-two on 19 of 20 blocks, and its worst-case error stays within 1.39$\times$ of the best
  result everywhere.
\end{enumerate}

% !TEX root = ../../SCDF.tex
\section{Related work}
\label{sec:related}

% Table: tool families and their embedded motion models.
\begin{table*}[t]
\centering\footnotesize
\setlength{\tabcolsep}{4pt}
\renewcommand{\arraystretch}{1.12}
\caption{Tool Families for Optical Co-Registration and Their Embedded
Motion Models}
\label{tab:families}
\begin{tabular}{@{}>{\raggedright\arraybackslash}p{0.23\textwidth}
                 >{\raggedright\arraybackslash}p{0.30\textwidth}
                 >{\raggedright\arraybackslash}p{0.185\textwidth}
                 >{\raggedright\arraybackslash}p{0.205\textwidth}@{}}
\toprule
Family & Motion model & Decision constants & Output \\
\midrule
Phase-correlation grids\newline
\emph{AROSICS; COSI-Corr}
  & translation per window, then a global shift or affine fit
  & thresholds tuned per product
  & tie-point grid $\to$ fitted model \\
\addlinespace
Dense chip tracking\newline
\emph{autoRIFT}
  & translation per chip, fixed search range
  & fixed ${\sim}25$\px{} window, quality thresholds
  & dense grid \\
\addlinespace
Optical flow\newline
\emph{GeFolki}
  & small, smooth flow
  & tuned (rank filter, radii)
  & dense per-pixel field \\
\addlinespace
Sparse features $+$ model\newline
\emph{SIFT+RANSAC; RIFT, RIFT2, SRIF}
  & single global affine or homography
  & ratio and inlier thresholds
  & sparse matches $\to$ one global transform \\
\addlinespace
Learned matchers\newline
\emph{LightGlue, LoFTR; RoMa}
  & learned prior; fixed working resolution, tiled at scene scale
  & learned weights, tuned acceptance
  & sparse/semi-dense matches ($\to$ fitted model) \\
\midrule
\method{} (ours)
  & none --- the dense per-pixel field \emph{is} the product
  & \textbf{self-calibrated per pair}
  & dense field $+$ confidence \\
\bottomrule
\end{tabular}
\end{table*}

Co-registration tooling splits into three families: area-based
estimators, sparse features fitted to a global model, and learned
matchers. Each embeds its assumptions as fixed constants somewhere,
whether a motion model, a search range, or an acceptance threshold. Table~\ref{tab:families}
summarizes those assumptions, and Section~\ref{sec:results-specialists}
later traces every catastrophic failure in the benchmark back to the
table's motion-model column. Evaluation practice is the fourth thread,
because on our data the way results are aggregated decides rankings
(Section~\ref{sec:benchmark-metrics}).

\paragraph{Area-based and dense methods}
AROSICS \cite{scheffler2017arosics} estimates tie points by
frequency-domain phase correlation on a regular grid and corrects with a
global shift or an affine fit to the validated tie points; the COSI-Corr lineage
\cite{leprince2007cosicorr} established sub-pixel image correlation for
deformation measurement. autoRIFT \cite{lei2021autorift} tracks dense
chips under a capped search window, demonstrated on smooth glacier-velocity
fields. GeFolki \cite{brigot2016gefolki} adapts Lucas--Kanade optical flow
to remote sensing with rank filtering. These tools are precise where their
assumptions hold; AROSICS takes more blocks outright than any other
baseline in our benchmark (Section~\ref{sec:results-specialists}). However, each
assumes translation per window or a small smooth flow, searches a fixed
range, and comes with thresholds tuned per product. None re-derives its
acceptance decisions from the pair at hand.

\paragraph{Sparse features with a global model}
SIFT with RANSAC \cite{lowe2004sift,fischler1981ransac} remains the
canonical sparse pipeline; for multimodal pairs, the RIFT family
\cite{li2020rift,li2023rift2} pairs phase-congruency detection with
log-Gabor index-map descriptors, and SRIF \cite{li2023srif} adds scale
invariance on a local intensity binary transform. Whatever the matcher, the product is one
global transform, which cannot represent non-global deformation: opposing
mosaic-seam shifts have no homography solution. Section~\ref{sec:results-specialists} quantifies the consequence on the
seam fixture, where the gap follows from the motion model and tuning does
not close it.

\paragraph{Learned matchers}
Learned sparse matching over SuperPoint features (SuperGlue, LightGlue)
\cite{detone2018superpoint,sarlin2020superglue,lindenberger2023lightglue},
detector-free matching (LoFTR and its successors)
\cite{sun2021loftr,wang2024efficientloftr}, and dense matchers (RoMa,
MASt3R) \cite{edstedt2024roma,leroy2024mast3r} dominate matching
benchmarks. The dense and semi-dense models run at small fixed working
resolutions and are memory-bound at scale; no published matcher
ingests an $8192^2$ scene natively, and practice at scene scale is tiled
inference \cite{li2025rsmatchsurvey,corley2026pretrained}. Like any trained
model, a learned matcher carries no performance guarantee outside its
training distribution \cite{shen2024gim,he2025matchanything}. This missing
guarantee, the operational burden, and the input-size limits together
motivate a training-free design.

\paragraph{Evaluation practice}
Remote-sensing registration papers conventionally report a checkpoint
root-mean-square error, a circular error at the 90th percentile (CE90), or
a mean end-point error over a handful of pairs. Matching benchmarks
elsewhere rank by outlier fractions under a fixed tolerance: Fl-all in
KITTI \cite{menze2015kitti}, mean matching accuracy on HPatches
\cite{dusmanu2019d2net,balntas2017hpatches}, pose-thresholded mean average
accuracy in the Image Matching Challenge \cite{jin2021imc}, and success
rates and matching precision in remote-sensing matching
\cite{li2020rift,ye2019fast}. None of this transfers unchanged to dense
co-registration fields. When pixel size varies widely across the pairs of
one benchmark, a fixed pixel tolerance means a different metric tolerance
on every pair; a single diverged extrapolation dominates a mean; and a
method that declines a hard pair drops out of its own average.
Sections~\ref{sec:benchmark-metrics}--\ref{sec:benchmark-stats} answer
these three failure modes. To our knowledge, no prior benchmark scores
dense co-registration fields of large optical scenes against constructed
dense truth on real pairs.

% !TEX root = ../../SCDF.tex
\section{Method}
\label{sec:method}

\subsection{Problem statement and conventions}
\label{sec:method-conventions}

Given a reference raster REF and a moving raster MOV, the deliverable is a
dense displacement field on \emph{MOV's own pixel grid},
\begin{equation}
  \mathbf{D}[y,x] \;=\; \mathbf{p}^{\mathrm{corrected}}_{\mathrm{geo}}(y,x)
                 \;-\; \mathbf{p}^{\mathrm{original}}_{\mathrm{geo}}(y,x)
  \;\in\; \mathbb{R}^2 ,
  \label{eq:dgeo}
\end{equation}
of shape $(H_{\mathrm{MOV}}, W_{\mathrm{MOV}}, 2)$, expressed in geographic
units in MOV's CRS. If the two CRSs differ, REF positions are converted to
MOV's CRS before differencing, so coordinate-system handling is contained in
$\mathbf{D}$ and does not recur downstream. A companion valid mask marks
the pixels where $\mathbf{D}$ is defined; it is undefined outside the
REF--MOV overlap and on nodata. Positions are anchored at pixel
centers throughout. Every method in the benchmark is scored on this
object and under this convention.

Internally the estimator carries two more same-shape companions: a flag map
$F$ marking which pixels of the valid region hold a directly measured match
(the rest are filled by interpolation), and a confidence map $C$, a
single lock-quality scale shared by every point source and used by the
interpolation weights, the non-maximum suppression, and the filters.

\subsection{Design principle: bounding the worst case}
\label{sec:method-principle}

A co-registration tool in production runs unattended on heterogeneous
inputs. The question is therefore not how precise a method is when
its assumptions hold, but whether some input regime makes it return a
field wrong by hundreds of meters, with nothing marking the failure. \method{} is designed to bound
that worst case, which reduces to refusing two kinds of commitment:

\begin{enumerate}
\item \emph{No global model.} The dense field is itself the motion model: it
      is never passed through a homography, an affine, or any other
      low-dimensional transform, so there is no global assumption for a
      scene to violate.
\item \emph{No tuned constant.} Every decision threshold is computed from
      the image pair being registered, at the point of use. With no absolute
      cutoffs to fit, the method is neither tuned for nor tuned against any
      particular regime.
\end{enumerate}

The design therefore makes a falsifiable prediction about variance across
conditions rather than about peak accuracy. Specialized for nothing, the
method should rarely be the single best on any one condition; but with no
assumption to violate and no threshold tuned to any regime, there should be
no condition on which its error degrades sharply.
Section~\ref{sec:results} tests this prediction; both halves hold.

\begin{figure*}[t]
  \centering
\begin{tikzpicture}[scdf]
\hyphenpenalty=10000 \exhyphenpenalty=10000

% --- geometry --------------------------------------------------------------
\def\ph{2.60cm}      % panel height, both bands
\def\pwA{3.35cm}     % panel width, band A
\def\pwB{4.60cm}     % panel width, band B
\def\capA{3.10cm}    % caption text width, band A
\def\capB{4.34cm}    % caption text width, band B

% band A, panel 1: the shared ground area (outside the loop, runs once)
\node[panel, minimum width=\pwA, minimum height=\ph, anchor=south west]
  (P1) at (0,0) {};
\node[stepnum] (s1) at ([shift={(0.30,-0.26)}]P1.north west) {1};
\node[paneltitle, anchor=west] at ([xshift=1.5mm]s1.east) {Shared ground};
\node[panelcap, text width=\capA, anchor=south] at ([yshift=1.3mm]P1.south)
  {the overlap, and one seeded point at its center};
\begin{scope}[shift={($(P1.center)+(0,0.12)$)}]
  \fill[scdfblue!8]   (-1.32,-0.18) rectangle (0.40,0.66);
  \fill[scdfgreen!12] (-0.40,-0.66) rectangle (1.32,0.18);
  \fill[scdfblue!32]  (-0.40,-0.18) rectangle (0.40,0.18);
  \draw[scdfblue,           line width=0.7pt] (-1.32,-0.18) rectangle (0.40,0.66);
  \draw[scdfgreen!62!black, line width=0.7pt] (-0.40,-0.66) rectangle (1.32,0.18);
  \node[minilbl, text=scdfblue,           anchor=north west] at (-1.29,0.64) {REF};
  \node[minilbl, text=scdfgreen!52!black, anchor=south east] at ( 1.29,-0.64) {MOV};
  \fill[scdfselfcal] (0,0) circle (2.0pt);
\end{scope}

% band A, panels 2-4: the per-scale loop.
% panel 2: grid on MOV, predicted window on REF
\node[panel, minimum width=\pwA, minimum height=\ph, anchor=south west]
  (P2) at (4.05,0) {};
\node[stepnum] (s2) at ([shift={(0.30,-0.26)}]P2.north west) {2};
\node[paneltitle, anchor=west] at ([xshift=1.5mm]s2.east) {Where to look};
\node[panelcap, text width=\capA, anchor=south] at ([yshift=1.3mm]P2.south)
  {the field so far predicts each cell's window in REF};
\begin{scope}[shift={($(P2.center)+(0,0.12)$)}]
  % MOV's share of the overlap, gridded, one cell picked out
  \fill[scdfgreen!10] (-1.42,-0.36) rectangle (-0.32,0.66);
  \foreach \i in {1,2}{%
    \draw[gridline] ({-1.42+\i*0.3667},-0.36) -- ({-1.42+\i*0.3667},0.66);
    \draw[gridline] (-1.42,{-0.36+\i*0.34}) -- (-0.32,{-0.36+\i*0.34});}
  \draw[scdfgreen!62!black, line width=0.6pt] (-1.42,-0.36) rectangle (-0.32,0.66);
  \draw[cellfill] (-1.0533,-0.02) rectangle (-0.6867,0.32);
  % REF, with the predicted window (solid) inside a search margin (dashed)
  \fill[scdfblue!7] (0.32,-0.36) rectangle (1.42,0.66);
  \draw[scdfblue, line width=0.6pt] (0.32,-0.36) rectangle (1.42,0.66);
  \draw[winbox] (0.62,-0.12) rectangle (1.18,0.44);
  \draw[scdfselfcal!75, line width=0.5pt] (0.72,-0.02) rectangle (1.08,0.34);
  % the prior that connects the two
  \draw[-{Stealth[length=3pt,width=2.4pt]}, draw=scdfselfcal, line width=0.55pt]
    (-0.62,0.18) to[out=20,in=160] (0.58,0.20);
  \node[minilbl, text=scdfselfcal, anchor=south] at (-0.02,0.28) {prior};
  \node[minilbl, text=scdfgreen!52!black, anchor=north] at (-0.87,-0.39) {MOV};
  \node[minilbl, text=scdfblue,           anchor=north] at ( 0.87,-0.39) {REF};
\end{scope}

% panel 3: the matcher, left closed
\node[panel, minimum width=\pwA, minimum height=\ph, anchor=south west]
  (P3) at (7.75,0) {};
\node[stepnum] (s3) at ([shift={(0.30,-0.26)}]P3.north west) {3};
\node[paneltitle, anchor=west] at ([xshift=1.5mm]s3.east) {Match};
\node[panelcap, text width=\capA, anchor=south] at ([yshift=1.3mm]P3.south)
  {any matcher --- keypoints, semi-dense, or dense};
\begin{scope}[shift={($(P3.center)+(0,0.12)$)}]
  \fill[scdfgreen!10] (-1.30,-0.40) rectangle (-0.50,0.44);
  \draw[scdfgreen!62!black, line width=0.6pt] (-1.30,-0.40) rectangle (-0.50,0.44);
  \fill[scdfblue!7] (0.50,-0.40) rectangle (1.30,0.44);
  \draw[scdfblue, line width=0.6pt] (0.50,-0.40) rectangle (1.30,0.44);
  \foreach \ax/\ay/\bx/\by in {%
      -1.10/0.22/0.70/0.26, -0.72/-0.04/1.08/0.00,
      -1.02/-0.26/0.78/-0.22, -0.64/0.32/1.16/0.36}{%
    \draw[draw=scdfink!45, line width=0.45pt] (\ax,\ay) -- (\bx,\by);
    \fill[scdfink!75] (\ax,\ay) circle (1.0pt);
    \fill[scdfink!75] (\bx,\by) circle (1.0pt);}
\end{scope}

% panel 4: the consistency filter, left closed
\node[panel, minimum width=\pwA, minimum height=\ph, anchor=south west]
  (P4) at (11.45,0) {};
\node[stepnum] (s4) at ([shift={(0.30,-0.26)}]P4.north west) {4};
\node[paneltitle, anchor=west] at ([xshift=1.5mm]s4.east) {Filter};
\node[panelcap, text width=\capA, anchor=south] at ([yshift=1.3mm]P4.south)
  {drop outliers --- against the whole set, then against their neighbors};
\begin{scope}[shift={($(P4.center)+(0,0.14)$)}]
  % a locally smooth displacement field, and three points that are not
  \foreach \x/\y/\a in {%
      -1.05/0.34/28, -0.55/0.34/24, -0.05/0.34/31, 0.95/0.34/29,
      -1.05/0.00/25,               -0.05/0.00/27, 0.45/0.00/24, 0.95/0.00/28,
      -1.05/-0.34/30, -0.55/-0.34/26, 0.45/-0.34/25, 0.95/-0.34/27}{%
    \draw[-{Stealth[length=2.2pt,width=1.8pt]}, draw=scdfblue!85,
          line width=0.5pt] (\x,\y) -- ++(\a:0.26);}
  % the long arrow fails the global magnitude gate; the two short
  % misdirected ones fail the neighbor test
  \foreach \x/\y/\a/\l in {0.45/0.34/232/0.26, -0.55/0.00/205/0.50,
                           -0.05/-0.34/118/0.26}{%
    \draw[-{Stealth[length=2.2pt,width=1.8pt]}, draw=scdfselfcal!55,
          line width=0.5pt] (\x,\y) -- ++(\a:\l);
    \draw[scdfselfcal, line width=0.75pt]
      ([shift={(-0.07,-0.07)}]\x,\y) -- ([shift={(0.07,0.07)}]\x,\y);
    \draw[scdfselfcal, line width=0.75pt]
      ([shift={(-0.07,0.07)}]\x,\y) -- ([shift={(0.07,-0.07)}]\x,\y);}
\end{scope}

% the loop group and its return path
\coordinate (lb4) at ([yshift=-0.46cm]P4.south);
\draw[looparr] (P4.south) -- (lb4) -- (lb4 -| P2.south) -- (P2.south);
% The scale strip, on the return path: one fixed-size plate per level, the
% patch a bold tile quartering in area from plate to plate. Keep the tiling
% grid faint, so it reads as "the patch shrinks", not "the grid densifies".
\coordinate (pyrC) at ($(lb4 -| P2.south)!0.72!(lb4)$);
\begin{scope}[shift={(pyrC)}]
  \fill[scdfstagefill] (-1.55,-0.33) rectangle (1.55,0.33);
  % level 1: the patch IS the whole overlap — and the first few points land
  \fill[scdfgreen!35] (0.85,-0.29) rectangle (1.43,0.29);
  \draw[scdfgreen!62!black, line width=0.55pt] (0.85,-0.29) rectangle (1.43,0.29);
  \foreach \x/\y/\a in {0.99/-0.11/36, 1.21/0.03/28}{%
    \draw[-{Stealth[length=1.7pt,width=1.5pt]}, draw=scdfblue!90,
          line width=0.42pt] (\x,\y) -- ++(\a:0.11);}
  % level 2: halved — one bold tile on a faintly 2x2-tiled plate
  \fill[scdfgreen!10] (-0.29,-0.29) rectangle (0.29,0.29);
  \draw[draw=scdfgreen!58!black!35, line width=0.35pt] (0,-0.29) -- (0,0.29);
  \draw[draw=scdfgreen!58!black!35, line width=0.35pt] (-0.29,0) -- (0.29,0);
  \fill[scdfgreen!35] (-0.29,0) rectangle (0,0.29);
  \draw[scdfgreen!62!black, line width=0.5pt] (-0.29,0) rectangle (0,0.29);
  \draw[scdfgreen!62!black, line width=0.55pt] (-0.29,-0.29) rectangle (0.29,0.29);
  \foreach \x/\y/\a in {-0.17/0.07/40, 0.10/0.12/30, -0.15/-0.19/34,
                        0.13/-0.13/26, -0.01/-0.03/44}{%
    \draw[-{Stealth[length=1.7pt,width=1.5pt]}, draw=scdfblue!90,
          line width=0.42pt] (\x,\y) -- ++(\a:0.10);}
  % level 3: halved again
  \fill[scdfgreen!10] (-1.43,-0.29) rectangle (-0.85,0.29);
  \foreach \k in {1,2,3}{%
    \draw[draw=scdfgreen!58!black!35, line width=0.3pt]
      ({-1.43+\k*0.145},-0.29) -- ({-1.43+\k*0.145},0.29);
    \draw[draw=scdfgreen!58!black!35, line width=0.3pt]
      (-1.43,{-0.29+\k*0.145}) -- (-0.85,{-0.29+\k*0.145});}
  \fill[scdfgreen!35] (-1.43,0.145) rectangle (-1.285,0.29);
  \draw[scdfgreen!62!black, line width=0.5pt] (-1.43,0.145) rectangle (-1.285,0.29);
  \draw[scdfgreen!62!black, line width=0.55pt] (-1.43,-0.29) rectangle (-0.85,0.29);
  \foreach \x/\y/\a in {-1.36/0.14/38, -1.18/0.18/28, -1.00/0.13/44,
                        -1.37/-0.03/30, -1.19/0.00/36, -1.00/-0.04/26,
                        -1.36/-0.20/42, -1.18/-0.17/32, -0.99/-0.21/38}{%
    \draw[-{Stealth[length=1.7pt,width=1.5pt]}, draw=scdfblue!90,
          line width=0.42pt] (\x,\y) -- ++(\a:0.09);}
  % scale-to-scale flow, in the return path's direction
  \draw[-{Stealth[length=2.4pt,width=2pt]}, draw=scdfblue!70!black,
        line width=0.5pt] (0.77,0) -- (0.37,0);
  \draw[-{Stealth[length=2.4pt,width=2pt]}, draw=scdfblue!70!black,
        line width=0.5pt] (-0.37,0) -- (-0.77,0);
\end{scope}
\coordinate (pyrB) at ($(pyrC)+(0,-0.35)$);
\node[font=\sffamily\tiny, text=scdfblue!70!black, fill=scdfstagefill,
      align=center, inner sep=2pt] at ($(lb4 -| P2.south)!0.26!(lb4)$)
  {next scale --- the patch halves;\\ only the field so far carries over};
\node[phasetitle, anchor=south west] (tA) at ([yshift=1.2mm]P2.north west)
  {Coarse-to-fine matching cascade};
\begin{pgfonlayer}{background}
  \node[phasebox, fit={(tA)(P2)(P4)(lb4)(pyrB)}, inner sep=6pt] (bandA) {};
\end{pgfonlayer}
\draw[flowarr] (P1.east) -- (P1.east -| bandA.west);

% band B, panels 5-7: once, after the pyramid.
% panel 5: the correlation pass
\node[panel, minimum width=\pwB, minimum height=\ph, anchor=north west]
  (P5) at (0.20,-2.25) {};
\node[stepnum] (s5) at ([shift={(0.30,-0.26)}]P5.north west) {5};
\node[paneltitle, anchor=west] at ([xshift=1.5mm]s5.east)
  {Re-measure, and densify};
\node[panelcap, text width=\capB, anchor=south] at ([yshift=1.3mm]P5.south)
  {each point re-measured inside its own pixel; a lattice fills the gaps};
\begin{scope}[shift={($(P5.center)+(0,0.12)$)}]
  % left: a pixel zoom (gray ink, big cells, not a field footprint) — the
  % correction stays inside one pixel
  \fill[scdfselfcal!10] (-1.53,-0.14) rectangle (-1.17,0.22);
  \foreach \k in {1,2}{%
    \draw[draw=scdfink!32, line width=0.45pt]
      ({-1.89+\k*0.36},-0.50) -- ({-1.89+\k*0.36},0.58);
    \draw[draw=scdfink!32, line width=0.45pt]
      (-1.89,{-0.50+\k*0.36}) -- (-0.81,{-0.50+\k*0.36});}
  \draw[draw=scdfink!55, line width=0.7pt] (-1.89,-0.50) rectangle (-0.81,0.58);
  \fill[scdfblue]    (-1.46,0.13) circle (1.6pt);
  \fill[scdfselfcal] (-1.26,-0.04) circle (1.6pt);
  \draw[-{Stealth[length=3pt,width=2.4pt]}, draw=scdfselfcal,
        line width=0.65pt] (-1.42,0.10) -- (-1.31,0.00);
  \draw[draw=scdfink!20, line width=0.4pt] (0,-0.44) -- (0,0.52);
  % right: the field plate, sparse feature points plus the lattice — the
  % same square as panel 6, since densification adds points, never extent
  \fill[scdfgreen!12] (0.81,-0.50) rectangle (1.89,0.58);
  \foreach \x in {1.08,1.35,1.62}{%
    \draw[draw=scdfselfcal!28, line width=0.35pt] (\x,-0.50) -- (\x,0.58);}
  \foreach \y in {-0.23,0.04,0.31}{%
    \draw[draw=scdfselfcal!28, line width=0.35pt] (0.81,\y) -- (1.89,\y);}
  \draw[scdfgreen!62!black, line width=0.7pt] (0.81,-0.50) rectangle (1.89,0.58);
  \foreach \x in {1.08,1.35,1.62}{%
    \foreach \y in {-0.23,0.04,0.31}{%
      \fill[scdfselfcal!85] (\x,\y) circle (1.1pt);}}
  \foreach \x/\y in {0.95/0.44, 1.46/0.28, 1.80/0.46, 1.12/-0.14, 1.68/-0.34}{%
    \fill[scdfblue] (\x,\y) circle (1.7pt);}
\end{scope}

% panel 6: the fill
\node[panel, minimum width=\pwB, minimum height=\ph, anchor=north west]
  (P6) at (5.20,-2.25) {};
\node[stepnum] (s6) at ([shift={(0.30,-0.26)}]P6.north west) {6};
\node[paneltitle, anchor=west] at ([xshift=1.5mm]s6.east) {Fill};
\node[panelcap, text width=\capB, anchor=south] at ([yshift=1.3mm]P6.south)
  {sparse vectors in, a dense displacement field out --- cross-validated
   interpolator};
\begin{scope}[shift={($(P6.center)+(0,0.12)$)}]
  % left: the accepted control points, on the same square as the dense side
  \draw[draw=scdfink!35, line width=0.5pt] (-1.89,-0.50) rectangle (-0.81,0.58);
  \foreach \x/\y in {-1.71/0.34, -1.33/0.42, -0.99/0.26, -1.55/0.10,
                     -1.11/-0.02, -1.73/-0.20, -1.35/-0.30, -1.01/-0.36}{%
    \pgfmathsetmacro{\ang}{20 + 25*(\x+1.89)/1.08 + 25*(\y+0.50)/1.08}%
    \fill[scdfblue] (\x,\y) circle (1.2pt);
    \draw[-{Stealth[length=2pt,width=1.7pt]}, draw=scdfblue, line width=0.5pt]
      (\x,\y) -- ++(\ang:0.15);}
  \draw[-{Stealth[length=3pt,width=2.4pt]}, draw=scdfink!70, line width=0.55pt]
    (-0.68,0.04) -- (0.68,0.04);
  % right: the same field, dense. This glyph is repeated stroke for stroke
  % as panel 7's input — edit both together.
  \fill[scdfblue!5] (0.81,-0.50) rectangle (1.89,0.58);
  \draw[draw=scdfink!35, line width=0.5pt] (0.81,-0.50) rectangle (1.89,0.58);
  \foreach \i in {0,...,5}{%
    \foreach \j in {0,...,5}{%
      \pgfmathsetmacro{\ang}{20 + 5*\i + 5*\j}%
      \draw[-{Stealth[length=1.6pt,width=1.4pt]}, draw=scdfblue!80,
            line width=0.4pt]
        ({0.93+\i*0.168},{-0.38+\j*0.168}) -- ++(\ang:0.10);}}
\end{scope}

% panel 7: the warp
\node[panel, minimum width=\pwB, minimum height=\ph, anchor=north west]
  (P7) at (10.20,-2.25) {};
\node[stepnum] (s7) at ([shift={(0.30,-0.26)}]P7.north west) {7};
\node[paneltitle, anchor=west] at ([xshift=1.5mm]s7.east) {Warp};
\node[panelcap, text width=\capB, anchor=south] at ([yshift=1.3mm]P7.south)
  {MOV resampled through the dense field --- identical for every method
   compared};
\begin{scope}[shift={($(P7.center)+(0,0.12)$)}]
  % in: the dense field from panel 6, the same quiver stroke for stroke
  \fill[scdfblue!5] (-1.89,-0.50) rectangle (-0.81,0.58);
  \draw[draw=scdfink!35, line width=0.5pt] (-1.89,-0.50) rectangle (-0.81,0.58);
  \foreach \i in {0,...,5}{%
    \foreach \j in {0,...,5}{%
      \pgfmathsetmacro{\ang}{20 + 5*\i + 5*\j}%
      \draw[-{Stealth[length=1.6pt,width=1.4pt]}, draw=scdfblue!80,
            line width=0.4pt]
        ({-1.77+\i*0.168},{-0.38+\j*0.168}) -- ++(\ang:0.10);}}
  \draw[-{Stealth[length=3pt,width=2.4pt]}, draw=scdfink!70, line width=0.55pt]
    (-0.68,0.04) -- (0.55,0.04);
  % out: MOV's grid carried onto REF — the one footprint in band B whose
  % shape differs, because warp resamples
  \fill[scdfblue!16] (0.68,-0.38) rectangle (2.10,0.42);
  \draw[scdfblue, line width=0.7pt] (0.68,-0.38) rectangle (2.10,0.42);
  % the moving grid, warped onto the reference
  \foreach \x in {1.03,1.39,1.75}{%
    \draw[gridline] (\x,-0.38) to[out=78,in=-102] ({\x+0.12},0.42);}
  \foreach \y in {-0.11,0.15}{%
    \draw[gridline] (0.68,\y) to[out=9,in=189] (2.10,{\y+0.09});}
\end{scope}

\node[phasetitle, anchor=south west] (tB) at ([yshift=1.2mm]P5.north west)
  {Post-processing};
\begin{pgfonlayer}{background}
  \node[phasebox, fit={(tB)(P5)(P7)}, inner sep=6pt] (bandB) {};
\end{pgfonlayer}
\draw[flowarr] (bandA.south) -- (bandA.south |- bandB.north);

\end{tikzpicture}
  \caption{How \method{} estimates the field: one seeded point \stepdot{1};
  a predict--measure--filter loop \stepdot{2}--\stepdot{4}, repeated coarse to
  fine with the patch halving at each scale and only the field carrying over;
  a correlation pass \stepdot{5} that re-measures every point and densifies
  the set; interpolation \stepdot{6} to a dense displacement field; and the
  warp \stepdot{7} that resamples MOV through it. Figure~\ref{fig:corrpass}
  details the measurement inside \stepdot{5}.}
  \label{fig:pipeline}
\end{figure*}
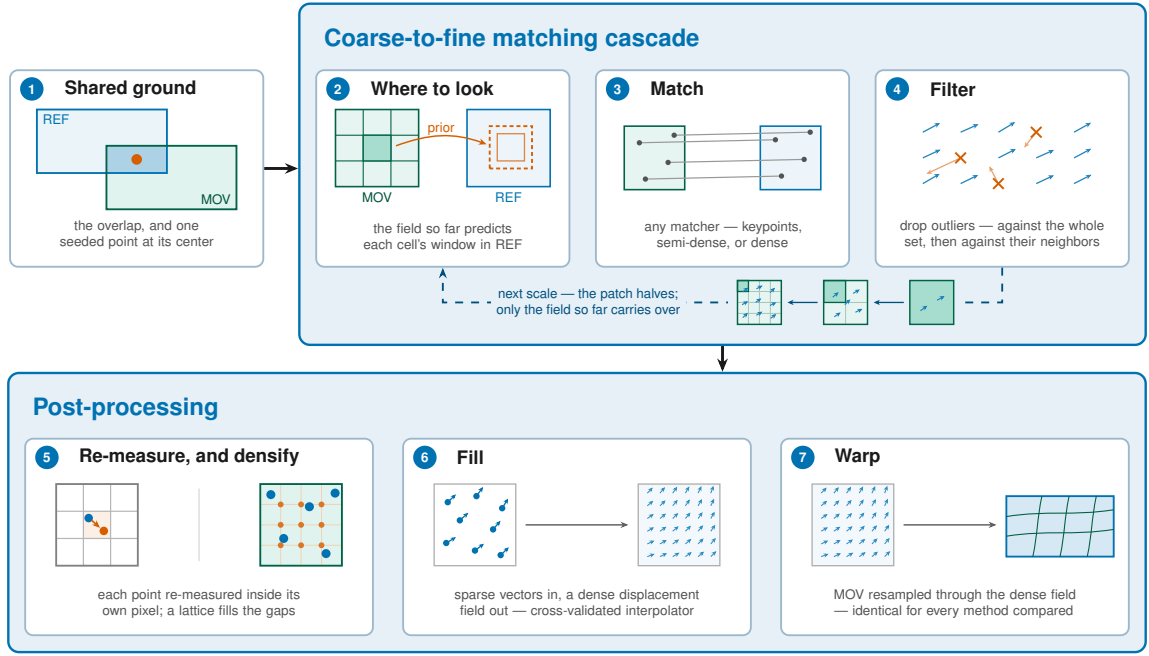

The remainder of the section follows Figure~\ref{fig:pipeline} stage by
stage. For each stage, Table~\ref{tab:selfcal} records where its decision
thresholds come from.

% !TEX root = ../../SCDF.tex
% Figure: the SCDF estimator, drawn rather than described. Band A is the
% per-scale loop (panel 1 stands outside it, running once); band B is the
% post-processing that runs once after the pyramid. Panels are a fixed
% 2.60 cm tall, with artwork drawn in a scope shifted to the panel center
% and confined to x in [-1.55, 1.55] (band A) / [-2.16, 2.16] (band B),
% y in [-0.66, 0.66].

\subsection{The cascade: one loop, repeated per level}
\label{sec:method-cascade}

The estimator is a single predict--measure--filter loop run over a
resolution pyramid. Patch size starts at the full REF/MOV geographic overlap
and halves per level down to a floor of 256\px; the field accumulated so far
is the only communication between levels.

\paragraph{Initialization}
The estimator computes the geographic overlap of the pair, estimates the
ground-sample-distance (GSD) ratio at its center, and seeds a single
zero-displacement point there. There is no exhaustive global search: every
later search window is predicted from the current field, and a large initial
offset is absorbed by the coarsest level, whose patch is the entire overlap.

\paragraph{Predict}
Patch centers are laid out uniformly on the MOV overlap grid, and at each
center a displacement prior is interpolated from all accumulated points by
moving least squares (MLS), a confidence-weighted local affine model
following Schaefer \emph{et al.} \cite{schaefer2006mls}. The prior,
together with a local GSD
ratio, places the REF search window with margin; when the resolutions
differ, the finer side is downsampled to the common grid.

\paragraph{Measure}
A pluggable patch matcher (RootSIFT by default
\cite{lowe2004sift,arandjelovic2012rootsift}) produces correspondences
that must survive a ratio test, a local RANSAC affine
\cite{fischler1981ransac}, and a forward/backward cycle-consistency check.
Each inlier becomes a displacement sample with a confidence. The most
confident candidate per pixel is committed to the estimator's state: its
displacement enters $\mathbf{D}$, the pixel is flagged as measured in $F$,
and its confidence is recorded in $C$; a stored match is overwritten
only by a more confident one.

\paragraph{Filter}
Reliability is enforced here, and at every level, so
that one level's mistakes cannot become the next level's prior. Write
$\{(\mathbf{p}_i, \mathbf{d}_i, c_i)\}$ for the level's accumulated points
(position on the MOV grid, displacement, confidence). Three tests run in
sequence. A global magnitude gate removes gross outliers first, keeping
point $i$ only if
\begin{equation}
  \lVert\mathbf{d}_i\rVert \;\le\;
  \mathop{\mathrm{med}}_{j} \lVert\mathbf{d}_j\rVert + 3\,\hat\sigma,
  \qquad
  \hat\sigma = 1.4826 \mathop{\mathrm{MAD}}_{j} \lVert\mathbf{d}_j\rVert .
  \label{eq:madgate}
\end{equation}
Throughout, $\mathrm{MAD}_j\, x_j = \mathrm{med}_j \lvert x_j -
\mathrm{med}_k\, x_k \rvert$ is the median absolute deviation, and $1.4826
\approx 1/\Phi^{-1}(3/4)$ ($\Phi$ the standard normal CDF) is the
constant that turns it into a standard-deviation estimate under Gaussian
noise \cite{rousseeuw1993alternatives}. Eq.~\eqref{eq:madgate} is thus a
$3\sigma$ rule built from statistics with a 50\% breakdown point: a scale
estimated by mean and standard deviation would itself be corrupted by the
outliers the gate is meant to remove.
The main test is a leave-one-out (LOO) consistency test. The question it
puts to every point is: \emph{had this measurement never been made, what
displacement would the surrounding matches predict here, and does the
measurement agree?} A true field is locally smooth, so a correct match is
predictable from the matches around it, while a false correspondence lands
at a value no smooth field through its neighbors can explain. The point's
own value is therefore withheld from the fit (the leave-one-out
construction), and the prediction is fitted
through its $k{=}12$ nearest neighbors $\mathcal{N}_i$, each weighted by
confidence and proximity, with the same weighted local affine model that
MLS interpolation uses:
\begin{equation}
  \begin{aligned}
  \hat{\mathbf{d}}_{-i} &= \mathbf{A}^{\!\star}\mathbf{p}_i +
  \mathbf{b}^{\star}, \\
  (\mathbf{A}^{\!\star}\!,\mathbf{b}^{\star}) &=
  \operatorname*{arg\,min}_{\mathbf{A},\,\mathbf{b}}
  \sum_{j \in \mathcal{N}_i}
  \frac{c_j}{\lVert\mathbf{p}_j - \mathbf{p}_i\rVert^{2}}
  \,\bigl\lVert \mathbf{A}\mathbf{p}_j + \mathbf{b} - \mathbf{d}_j
  \bigr\rVert^{2} .
  \end{aligned}
  \label{eq:loo}
\end{equation}
An affine model is used instead of a neighbor average because real fields
carry gradients (rotation, scale difference, terrain-induced shear) along
which neighboring displacements differ linearly. An average, like
any magnitude or angle heuristic, flags those points, whereas the
affine fit follows the trend, so a point on a smooth gradient matches its
prediction and only a point inconsistent with its own neighborhood fails.
The verdict compares the point's residual
$r_i = \lVert \mathbf{d}_i - \hat{\mathbf{d}}_{-i} \rVert$ against the
residuals of all points at this level,
\begin{equation}
  r_i \;\le\; \mathop{\mathrm{med}}_{j} r_j + 3.5\,\hat\sigma_r ,
  \qquad
  \hat\sigma_r = 1.4826 \mathop{\mathrm{MAD}}_{j}\, r_j ,
  \label{eq:loogate}
\end{equation}
the same robust construction as Eq.~\eqref{eq:madgate}, now calibrated on
this level's own distribution of disagreement rather than on
magnitudes. (Two refinements, both detailed in \supp{I}: residuals
are studentized for the prediction's leverage, so sparse hull-edge
neighborhoods are not over-penalized, and one Tukey-biweight pass reweights
the fits so an outlying neighbor cannot corrupt the verdicts of the points
around it.) Last, a deliberately small non-maximum suppression
(radius~32\px) thins duplicates, keeping the highest-confidence point per
neighborhood. The pipeline is \emph{recall-first} throughout: produce
many candidates, then delete with tests
calibrated on the pair.

% Table: every decision threshold in the pipeline and what it is read from.
% Italics in the right column mark the pair's-own sources.
\begin{table*}[t]
  \centering
  \caption{Self-Calibration Audit: Every Decision Threshold and Its Source}
  \label{tab:selfcal}
  \small
  \setlength{\tabcolsep}{5pt}
  \begin{tabular}{@{}>{\raggedright\arraybackslash}p{6.4cm}
                    >{\raggedright\arraybackslash}p{9.2cm}@{}}
    \toprule
    \textbf{Stage --- the decision it makes}
      & \textbf{Where its threshold comes from} \\
    \midrule
    \textbf{Global magnitude filter} --- is this displacement grossly out of
    family?
      & $\mathrm{med} + 3\,\hat\sigma$ of the \emph{displacement magnitudes
        at this level}, $\hat\sigma = 1.4826\times$MAD \\
    \addlinespace[3pt]
    \textbf{LOO local-affine filter} --- is this point inconsistent with its
    own 12 neighbors?
      & $\mathrm{med} + 3.5\,\hat\sigma$ of \emph{this level's own
        studentized residual distribution} \\
    \addlinespace[3pt]
    \textbf{Response gate} --- did this window pair lock, or is the peak
    spurious?
      & the maximum of $K{=}8$ \emph{null} correlations of the same REF
        window against far-away MOV windows, recomputed for every window \\
    \addlinespace[3pt]
    \textbf{Error gate} --- do the aligned windows agree well enough for
    \emph{this} pair?
      & a quantile of \emph{the pair's own error distribution} ($q{=}0.5$
        re-measuring, $0.4$ densifying)\textsuperscript{1} \\
    \addlinespace[3pt]
    \textbf{Fill interpolator} --- which of 7 interpolators suits this scene?
      & 5-fold cross-validation over \emph{the pair's own control points} \\
    \midrule
    \multicolumn{2}{@{}l@{}}{\emph{Fixed by design, and not thresholds on
      measured quantities:}} \\
    \addlinespace[2pt]
    \textbf{Match acceptance} --- ratio test and cycle check
      & Lowe ratio 0.75, forward/backward error $\le 3$\px{} --- front-end
        constants; the calibrated filters above carry the accuracy claims \\
    \addlinespace[3pt]
    \textbf{Search geometry} --- pyramid, windows, lattice
      & patch $\to$ 256\px{} floor, windows 128/64\px, lattice spacing 48,
        upsample $\times 32$, downsample cap 2048\px \\
    \addlinespace[3pt]
    \textbf{Prior-deviation bound} --- how far may a measurement move from
    the prediction?
      & 4 coarse px re-measuring, 3 densifying\textsuperscript{2} \\
    \bottomrule
    \addlinespace[3pt]
    \multicolumn{2}{@{}>{\raggedright\arraybackslash}
        p{\dimexpr 6.4cm+9.2cm+2\tabcolsep\relax}@{}}{\footnotesize
      \textsuperscript{1}~Clamped to $[0.6, 0.97]$ as a sanity interval; the
      clamp binds only when the pair's own quantile falls outside it.
      \textsuperscript{2}~An absolute bound, and the one place where a scale
      constant enters the loop: it prevents a single bad measurement from
      moving a point far from the prior the levels above established.} \\
  \end{tabular}
\end{table*}

\subsection{The correlation pass: re-measure and densify}
\label{sec:method-subpix}

The cascade ends with feature matches, only as precise as keypoint
localization: 0.3--0.5\px{} on the coarser of the two grids, multiplied on cross-resolution pairs by the resolution
ratio --- at the benchmark's most extreme pairing (1:16.7), half a
coarse pixel is already ${\sim}8$\px{} in the product. Features also leave gaps --- mosaic seams, scene
borders --- that the fill could only interpolate across. So after the
finest level the estimator runs its predict--measure--filter loop once
more, with the feature matcher replaced by direct, translation-only
correlation \cite{guizar2008subpixel}.

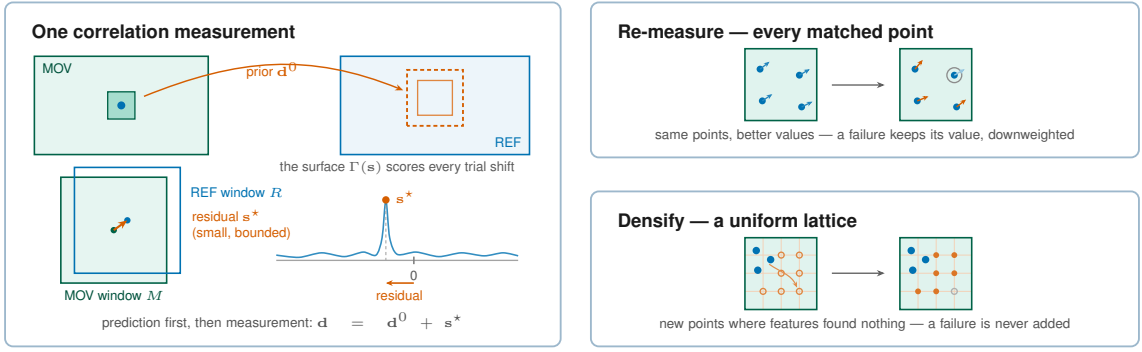
\begin{figure*}[t]
  \centering
\begin{tikzpicture}[scdf]
\hyphenpenalty=10000 \exhyphenpenalty=10000

% panel A: the measurement primitive
\node[panel, minimum width=7.35cm, minimum height=4.55cm, anchor=south west]
  (PA) at (0,0) {};
\node[paneltitle, anchor=north west] at ([shift={(0.25,-0.18)}]PA.north west)
  {One correlation measurement};
\node[panelcap, text width=7.0cm, anchor=south] at ([yshift=1.2mm]PA.south)
  {prediction first, then measurement:
   $\mathbf{d} = \mathbf{d}^{0} + \mathbf{s}^{\star}$};
% predict: the field places the REF window
\fill[scdfgreen!10] (0.40,2.55) rectangle (2.70,3.85);
\draw[scdfgreen!62!black, line width=0.6pt] (0.40,2.55) rectangle (2.70,3.85);
\draw[cellfill] (1.37,3.02) rectangle (1.73,3.38);
\fill[scdfblue] (1.55,3.20) circle (1.5pt);
\fill[scdfblue!7] (4.45,2.55) rectangle (6.95,3.85);
\draw[scdfblue, line width=0.6pt] (4.45,2.55) rectangle (6.95,3.85);
\draw[winbox] (5.33,2.93) rectangle (6.07,3.67);
\draw[scdfselfcal!75, line width=0.5pt] (5.47,3.07) rectangle (5.93,3.53);
\draw[-{Stealth[length=3pt,width=2.4pt]}, draw=scdfselfcal, line width=0.55pt]
  (1.85,3.32) to[out=22,in=158] (5.28,3.40);
\node[minilbl, text=scdfselfcal, anchor=south] at (3.55,3.50)
  {prior $\mathbf{d}^{0}$};
\node[minilbl, text=scdfgreen!52!black, anchor=north west] at (0.46,3.81) {MOV};
\node[minilbl, text=scdfblue, anchor=south east] at (6.89,2.60) {REF};
% the extracted pair: almost aligned, a small residual left
\fill[scdfgreen!10] (0.75,0.85) rectangle (2.15,2.25);
\draw[scdfgreen!62!black, line width=0.6pt] (0.75,0.85) rectangle (2.15,2.25);
\draw[scdfblue, line width=0.6pt] (0.93,0.98) rectangle (2.33,2.38);
\fill[scdfgreen!62!black] (1.45,1.55) circle (1.2pt);
\fill[scdfblue] (1.63,1.68) circle (1.2pt);
\draw[-{Stealth[length=3.5pt,width=2.8pt]}, draw=scdfselfcal, line width=0.8pt]
  (1.45,1.55) -- (1.63,1.68);
\node[minilbl, text=scdfselfcal, anchor=west, align=left] at (2.44,1.62)
  {residual $\mathbf{s}^{\star}$\\ (small, bounded)};
\node[minilbl, text=scdfgreen!52!black, anchor=north west] at (0.75,0.80)
  {MOV window $M$};
\node[minilbl, text=scdfblue, anchor=west] at (2.42,2.05) {REF window $R$};
% the correlation surface reads the residual, sub-pixel
\node[minilbl] at (5.20,2.38)
  {the surface $\Gamma(\mathbf{s})$ scores every trial shift};
\draw[draw=scdfink!40, line width=0.5pt] (3.60,1.15) -- (6.80,1.15);
\draw[draw=scdfblue!80, line width=0.6pt] plot[smooth, tension=0.5]
  coordinates {(3.60,1.22) (3.90,1.19) (4.20,1.25) (4.50,1.20) (4.75,1.26)
    (4.92,1.22) (5.00,1.45) (5.05,1.95) (5.10,1.45) (5.20,1.23) (5.45,1.26)
    (5.75,1.20) (6.05,1.25) (6.35,1.21) (6.60,1.25) (6.80,1.21)};
\fill[scdfselfcal] (5.05,1.95) circle (1.4pt);
\node[minilbl, text=scdfselfcal, anchor=west] at (5.14,1.99)
  {$\mathbf{s}^{\star}$};
\draw[draw=scdfink!40, line width=0.4pt, dash pattern=on 1.2pt off 1.0pt]
  (5.05,1.88) -- (5.05,1.15);
\draw[draw=scdfink!55, line width=0.6pt] (5.42,1.10) -- (5.42,1.20);
\node[minilbl, anchor=north] at (5.42,1.06) {0};
\draw[-{Stealth[length=2.6pt,width=2.2pt]}, draw=scdfselfcal, line width=0.6pt]
  (5.42,0.85) -- (5.05,0.85);
\node[minilbl, text=scdfselfcal, anchor=north] at (5.23,0.80) {residual};

% panel B: re-measure, the primitive at every matched point
\node[panel, minimum width=7.25cm, minimum height=2.05cm, anchor=south west]
  (PB) at (7.75,2.50) {};
\node[paneltitle, anchor=north west] at ([shift={(0.25,-0.18)}]PB.north west)
  {Re-measure --- every matched point};
\node[panelcap, text width=6.9cm, anchor=south] at ([yshift=1.0mm]PB.south)
  {same points, better values --- a failure keeps its value, downweighted};
\fill[scdfgreen!12] (9.80,3.00) rectangle (10.75,3.95);
\draw[scdfgreen!62!black, line width=0.6pt] (9.80,3.00) rectangle (10.75,3.95);
\foreach \x/\y/\a in {10.00/3.68/38, 10.52/3.60/30, 10.03/3.26/34,
                      10.55/3.18/26}{%
  \fill[scdfblue] (\x,\y) circle (1.3pt);
  \draw[-{Stealth[length=2.2pt,width=1.8pt]}, draw=scdfblue!85,
        line width=0.5pt] (\x,\y) -- ++(\a:0.16);}
\draw[thinarr] (10.95,3.47) -- (11.65,3.47);
\fill[scdfgreen!12] (11.85,3.00) rectangle (12.80,3.95);
\draw[scdfgreen!62!black, line width=0.6pt] (11.85,3.00) rectangle (12.80,3.95);
% three values replaced; one point failed the gates and keeps its value,
% faded and ringed to mark the reduced weight
\foreach \x/\y/\a in {12.05/3.68/50, 12.08/3.26/22, 12.60/3.18/38}{%
  \fill[scdfblue] (\x,\y) circle (1.3pt);
  \draw[-{Stealth[length=2.2pt,width=1.8pt]}, draw=scdfselfcal,
        line width=0.55pt] (\x,\y) -- ++(\a:0.16);}
\fill[scdfblue] (12.57,3.60) circle (1.3pt);
\draw[scdfgrey, line width=0.5pt] (12.57,3.60) circle (2.8pt);
\draw[-{Stealth[length=2.2pt,width=1.8pt]}, draw=scdfblue!40,
      line width=0.5pt] (12.57,3.60) -- ++(30:0.16);

% panel C: densify, the primitive on a uniform lattice
\node[panel, minimum width=7.25cm, minimum height=2.05cm, anchor=south west]
  (PC) at (7.75,0) {};
\node[paneltitle, anchor=north west] at ([shift={(0.25,-0.18)}]PC.north west)
  {Densify --- a uniform lattice};
\node[panelcap, text width=6.9cm, anchor=south] at ([yshift=1.0mm]PC.south)
  {new points where features found nothing --- a failure is never added};
\fill[scdfgreen!12] (9.80,0.50) rectangle (10.75,1.45);
\foreach \x in {10.04,10.28,10.52}{%
  \draw[draw=scdfselfcal!28, line width=0.35pt] (\x,0.50) -- (\x,1.45);}
\foreach \y in {0.74,0.98,1.22}{%
  \draw[draw=scdfselfcal!28, line width=0.35pt] (9.80,\y) -- (10.75,\y);}
\draw[scdfgreen!62!black, line width=0.6pt] (9.80,0.50) rectangle (10.75,1.45);
% candidates (hollow) skip cells that already hold feature points
\foreach \x/\y in {10.28/1.22, 10.52/1.22, 10.28/0.98, 10.52/0.98,
                   10.04/0.74, 10.28/0.74, 10.52/0.74}{%
  \draw[scdfselfcal!75, line width=0.5pt] (\x,\y) circle (1.1pt);}
\foreach \x/\y in {9.95/1.28, 10.14/1.16, 9.97/1.02}{%
  \fill[scdfblue] (\x,\y) circle (1.4pt);}
\draw[-{Stealth[length=2.2pt,width=1.8pt]}, draw=scdfselfcal!70,
      line width=0.45pt] (10.12,1.10) to[out=-35,in=115] (10.49,0.79);
\draw[thinarr] (10.95,0.98) -- (11.65,0.98);
\fill[scdfgreen!12] (11.85,0.50) rectangle (12.80,1.45);
\foreach \x in {12.09,12.33,12.57}{%
  \draw[draw=scdfselfcal!28, line width=0.35pt] (\x,0.50) -- (\x,1.45);}
\foreach \y in {0.74,0.98,1.22}{%
  \draw[draw=scdfselfcal!28, line width=0.35pt] (11.85,\y) -- (12.80,\y);}
\draw[scdfgreen!62!black, line width=0.6pt] (11.85,0.50) rectangle (12.80,1.45);
% survivors become points; one candidate failed and simply is not there
\foreach \x/\y in {12.33/1.22, 12.57/1.22, 12.33/0.98, 12.57/0.98,
                   12.09/0.74, 12.33/0.74}{%
  \fill[scdfselfcal!85] (\x,\y) circle (1.1pt);}
\draw[scdfgrey!70, line width=0.5pt] (12.57,0.74) circle (1.1pt);
\foreach \x/\y in {12.00/1.28, 12.19/1.16, 12.02/1.02}{%
  \fill[scdfblue] (\x,\y) circle (1.4pt);}

\end{tikzpicture}
  \caption{The correlation pass. Left: one measurement --- the
  accumulated field predicts where each MOV window sits in REF, the
  correlation surface $\Gamma(\mathbf{s})$ scores every trial shift at
  once, and its peak $\mathbf{s}^{\star}$ is the sub-pixel residual,
  added to and bounded by the prior. Right: the same primitive used
  twice --- \emph{re-measure} replaces the value at every matched point,
  \emph{densify} adds surviving points on a fresh lattice; what fails the
  acceptance gates (\supp{I}) keeps its old
  value or is never added.}
  \label{fig:corrpass}
\end{figure*}

\paragraph{Re-measure, and densify}
\emph{Re-measuring} visits every matched point and estimates a
correction to it with one correlation measurement, defined below, the
point's own value serving as the prediction (Figure~\ref{fig:corrpass}).
Where the measurement is accepted, its corrected displacement replaces
the feature matcher's value. Where it is rejected, the point keeps its
displacement at a confidence scaled by its response (clip rule in
\supp{I}) ---
downweighted rather than deleted, because deletion would remove the
anchors at displacement discontinuities such as mosaic seams, the
structure a global model cannot represent. \emph{Densifying} is the same measurement applied to new points: it
visits a uniform lattice (spacing 48\px{}, cells that already hold a
match skipped), with the MLS interpolant of the neighbors as the
prediction and tighter constants (Table~\ref{tab:selfcal}), and adds a
point wherever the measurement is accepted. A lattice candidate whose
measurement is rejected is never added.

% !TEX root = ../../SCDF.tex
% Figure: the correlation pass drawn mechanically — one measurement
% primitive (panel A) applied to two point sources (panels B and C). Shares
% fig-pipeline's visual vocabulary, and its notation matches §3.4's prose.

\paragraph{One measurement}
Re-measuring and densifying call the same primitive: given a point
$\mathbf{p}$ and a predicted displacement $\mathbf{d}^{0}$, it returns
a corrected displacement and a verdict on whether to trust it. The
prediction supplies the alignment: one window is cut from MOV around
$\mathbf{p}$, one from REF where $\mathbf{d}^{0}$ says the same ground
lies, and were the prediction exact the two would coincide. The offset
that remains, the \emph{residual}, is what the measurement finds: the
correlation surface $\Gamma(\mathbf{s})$ scores every trial shift
$\mathbf{s}$ at once, its peak $\mathbf{s}^{\star}$ is the residual,
and the corrected displacement is $\mathbf{d}^{0} + \mathbf{s}^{\star}$.
Every correlation surface has a maximum, related windows or not, so the
peak is evidence only if it passes three acceptance gates (definitions
in \supp{I}). The \emph{response gate} asks whether the peak is sharp
--- a lone spike on a flat surface rather than the tallest of many
similar waves --- and sets the standard per window: the peak must beat
every response the same REF window achieves against $K{=}8$ far-away
MOV windows, guaranteed mismatches. The \emph{error gate} asks whether
the aligned windows look alike, a separate question (water agrees
without locking; cross-sensor pairs lock without agreeing), with a
threshold read from the batch's own error distribution. The
\emph{prior-deviation bound} (Table~\ref{tab:selfcal}) rejects a
residual of more than a few coarse pixels, so that no single peak can
move a point far from the field around it. An accepted point takes its
response as its confidence, so feature points, lattice points and
future learned matchers share one lock-quality scale in $C$.
Section~\ref{sec:results-components} scores the response gate against
ground truth.

The merged set then passes the global and LOO filters of
Section~\ref{sec:method-cascade} once more.

\subsection{Fill by cross-validated interpolation}
\label{sec:method-fill}

The remaining unmeasured pixels are filled by interpolating the control
points, and no single interpolator can be expected to dominate across
scenes: thin-plate splines \cite{bookstein1989tps} suit dense elastic
fields, low-$\alpha$ MLS structured ones, inverse-distance weighting
sparse noisy sets.
So this choice, too, is refused as a constant and is cross-validated on
the pair's own control points. The points are split at random into
five folds, one split shared by all seven candidates (an MLS $k/\alpha$
family, IDW, TPS), and each candidate predicts every point from the four
folds that exclude it. The score is the mean Euclidean error of these
held-out predictions against the measured displacements --- no ground
truth is consulted. The winner refits all points and fills the field.

\subsection{Warp}
\label{sec:method-warp}

The field is the product; warping is downstream of it, with no estimation
inside. The corrected coordinates $x_{\mathrm{geo}} + \mathbf{D}[\ldots,0]$
and $y_{\mathrm{geo}} + \mathbf{D}[\ldots,1]$ are attached to MOV as GDAL
geolocation arrays, and \texttt{gdal.Warp} resamples through them,
nearest-neighbor by default, leaving radiometry untouched. Every benchmark method,
ours and the baselines alike, is rendered through this same call, so warp
quality never confounds the comparison.

% !TEX root = ../../SCDF.tex
\section{Benchmark and evaluation protocol}
\label{sec:benchmark}

The benchmark is built from real satellite imagery: Sentinel-2, Landsat-8/9,
and NAIP acquisitions over 60 sites sampled worldwide, assembled into six
groups of image pairs that range from a same-scene pair to a five-year
temporal gap and a 1:16.7 resolution ratio. Real pairs come with no dense
displacement truth, so we inject known distortions into these pairs and
score each method against the injected field. The result is 584 constructed
pairs (\emph{cells}), each run by ten methods, giving 5840 runs whose
artifacts are all frozen and re-scorable.

\subsection{Sites and image pairs}
\label{sec:benchmark-legs}

% Table: benchmark composition (legs, sites, cells).
\begin{table*}[t]
\centering\small
\caption{Benchmark Composition: The Six Legs and Their 584 Cells}
\label{tab:legs}
\setlength{\tabcolsep}{4pt}
\begin{tabular}{@{}llllrrrrr@{}}
\toprule
& & & & & & & \multicolumn{2}{c}{MOV px (m)} \\
\cmidrule(l){8-9}
Leg & REF & MOV (distorted) & GSD & $\Delta t$ & Sites & Cells
    & A1/2/4 & A3 \\
\midrule
E1/b1 & S2 2024-07, 10\,m & same scene & 1:1 & --- & 25 & 100 & 10 & 30 \\
\midrule
E2/b1 & S2 2024-07 & S2 2024-01 (cross-season) & 1:1 & ${\sim}0.5$\,yr & 25 & 100 & 10 & 30 \\
E2/b2 & S2 2024-07 & S2 2023-07 & 1:1 & ${\sim}1$\,yr & 25 & 100 & 10 & 30 \\
E2/b3 & S2 2024-07 & S2 2019-07 & 1:1 & ${\sim}5$\,yr & 25 & 100 & 10 & 30 \\
E2/b4 & S2 10\,m & Landsat-8/9 30\,m & 1:3 & $\le 50$\,d & 25 & 100 & 30 & 90 \\
E2/b5 & S2 10\,m & NAIP 0.6\,m & 1:16.7 & 1--2\,yr & 21 & 84 & 0.6 & 1.8 \\
\midrule
Total & & & & & 146 & \textbf{584} & & \\
\bottomrule
\addlinespace[2pt]
\multicolumn{9}{@{}l@{}}{\footnotesize MOV px (m): the meter size of one
  MOV pixel, under A1/A2/A4 and under A3 (which coarsens MOV
  $\times 3$).} \\
\end{tabular}
\end{table*}

All imagery is sourced automatically from Microsoft Planetary Computer
\cite{microsoft2022planetary} and used at original digital numbers, with no
radiometric normalization anywhere in the pipeline. Scenes are $8192^2$
crops at the finer sensor's GSD (67\,Mpx); the Landsat half of leg b4 is
$2731^2$ at its native 30\,m.

\begin{figure}[t]
  \centering
  \includegraphics[width=\linewidth]{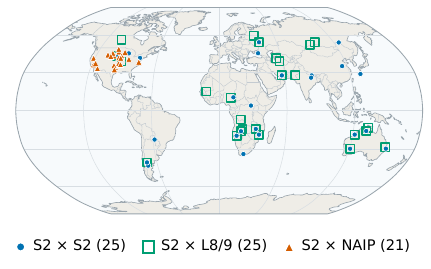}
  \caption{The 60 benchmark sites, marked by the image pairs acquired over
  them. A site shared between the Sentinel-2 and Landsat legs shows as a
  dot inside a ring; NAIP is a US program, so its sites are CONUS-only.}
  \label{fig:sites}
\end{figure}

Sites (Figure~\ref{fig:sites}) are drawn by global random sampling under two
acceptance rules on ESA WorldCover 2021 composition
\cite{zanaga2022worldcover}: an \emph{anti-homogeneity} gate (largest class
at most 90\% of the crop, at least two classes above 2\%) that rejects only
fully monotonous scenes such as open desert or ice sheets, and
a non-overlap rule keeping crop centers at least one footprint apart.
Weak-texture sites (water-dominated, semi-arid) are deliberately kept,
because they are where matchers fail in practice and sampling them away
would improve every method's apparent accuracy. Same-sensor pairs additionally pass per-crop
cloud/nodata quality control (thresholds in \supp{IV}).

% Figure: global site map.

Table~\ref{tab:legs} lists the six pair groups (\emph{legs}). One leg
distorts a single source --- a scene scored against its own distorted copy
(experiment E1, defined next). Three legs form a temporal-realism gradient
of real Sentinel-2 pairs, the same site revisited across half a year
(summer--winter), one year, and five years; every pair shares
its \emph{relative orbit}, enforced at fetch time. The last two legs are
cross-sensor: Sentinel-2 against Landsat-8/9 acquired within 30 days (1:3
GSD ratio), and against 0.6\,m NAIP aerial imagery within 15 days
(1:16.7).

Throughout the paper, one \emph{cell} is one (experiment, leg,
construction, site), a single constructed pair; one \emph{block} is one
(experiment, leg, construction), the 21--25 sites that share a fixture.
There are 584 cells in 24 blocks, scored by ten methods in
Section~\ref{sec:results}.

\subsection{Constructed ground truth on real pairs}
\label{sec:benchmark-construction}

A pair to be co-registered differs from its reference by more than the
displacement to be recovered. Acquisitions months or years apart see a
changed surface (season, vegetation, water); a different sensor brings
its own radiometry, blur, and noise; a different platform changes
viewing angle and pixel size; upstream processing leaves artifacts of
its own, mosaic seams among them. Real pairs, however, carry no
per-pixel displacement truth. The benchmark therefore injects a known
displacement field into a real scene and scores predictions against
the injected field, so that the truth is known exactly while the
imagery, with all the differences above, stays real. The four families of
Section~\ref{sec:benchmark-distortions} mirror those differences; they
do not follow a single motion model, as homography benchmarks do
\cite{balntas2017hpatches}.

Two experiment types share the four families and differ in the base
pair. \textbf{E1} builds MOV from REF itself: the pair differs by
exactly the injected distortion, so the ground truth, the
entire REF$\leftrightarrow$MOV displacement, is exact and gives an
uncontaminated floor. \textbf{E2} distorts the moving half of a
real pair: the genuine differences stay in the imagery, and the
truth is only the \emph{added} field. Its five legs sweep these
differences (cross-season to five-year revisit gaps, resolution
gaps of 1:3 and 1:16.7, spaceborne against airborne;
Table~\ref{tab:legs}), so E2 is the closer model of deployment.

The price of realism is the pair's pre-existing misalignment $r_0$,
which is not part of the constructed truth and floors the accuracy E2
can certify. The floor is identical for every method, so the
comparison stays fair. Literature documents the floor on four of the five legs:
same-relative-orbit Sentinel-2 pairs average 0.2--0.4\px{}
misregistration \cite{skakun2017coregistration,yan2018sentinel}
($r_0 \approx 3$--5\,m on b1--b3), and Landsat Collection-2 is
harmonized to the Sentinel-2 Global Reference Image to below 6\,m CE90
\cite{rengarajan2024coregistration}, small against b4's 30\,m pixel.
On b5 the floor is undocumented, so it supports \emph{ranking} only.
E1 and E2 are never pooled;
Section~\ref{sec:results-scope} shows they answer different questions.

\subsection{The four distortion families}
\label{sec:benchmark-distortions}

Each construction isolates, in controlled form, one of the differences
a real pair accumulates; together they range from geometry a single
global model absorbs exactly (A2) to geometry none fits (A4).

\begin{itemize}
\item \textbf{A1 --- sensor (radiometry and PSF):} each band is
tone-remapped on its own raw value scale (a $\gamma = 1.6$ curve with
random per-band gain and bias), blurred by a $\sigma = 2$\px{} Gaussian
emulating an MTF difference, and re-noised at 4\% of the band scale.
Geometry is untouched, so the truth is exactly zero.
\item \textbf{A2 --- viewing angle (homography):} the perspective
change of an off-nadir acquisition, applied as a projective warp that
moves the four scene corners by i.i.d.\ uniform offsets of up to
$0.02\,N$ ($\approx 164$\px).
\item \textbf{A3 --- resolution:} the grid change of a cross-GSD pair.
The mover is the nodata-aware area average of the base onto a
$3\times$ coarser grid at the same geolocation, so the truth is exactly
zero on the new grid.
\item \textbf{A4 --- non-global deformation (seam):} the discontinuity
of a real mosaic, in which the left and right halves shift in opposite
directions by $0.006\,N$ ($\pm 49$\px) across a smooth transition band,
plus a weak smooth elastic residual (${\sim}0.4$\px).
Seam-dominated by design, it admits no single global affine or
homography solution.
\end{itemize}

A construction is applied to the base raster (the source scene in
E1, the real mover in E2) as a pixel-space displacement field with
bilinear resampling; the injected field, expressed in map units on the
mover's own grid, is the ground truth (the convention of
Eq.~\eqref{eq:dgeo}). Magnitudes are fixed fractions of the scene side
$N$ ($N = 8192$ in the pixel figures above), one seed drives all
random draws, and the reference is never touched. We verify ground-truth
exactness by checking that the stored truth reproduces each
constructed mover exactly. Equations, parameters, the validity rule,
and the verification procedure are in \supp{III}.

\subsection{The scoring rule}
\label{sec:benchmark-metrics}

The rule by which the frozen matrix is condensed into a comparison changes
the winner on this data, consistent with
registration evaluation elsewhere, where protocol choices alone shift
measured error by up to $33\times$ for a fixed matcher
\cite{corley2026pretrained}. We therefore state the rule; each
clause repairs a specific way the obvious choice gives a wrong answer.

Within one cell we compute the per-pixel endpoint error (EPE)
$\|\mathbf{D}_{\mathrm{pred}} - \mathbf{D}_{\mathrm{gt}}\|$
\cite{baker2011flow} over valid pixels and summarize it by its median and
90th percentile (p90). Write $e_m(c)$ for method $m$'s per-cell median. Across cells, a block score is
\begin{equation}
  \begin{aligned}
  E_m(B) &\;=\; \operatorname*{median}_{c \in B}\; e_m(c), \\
  e_m(c) &= +\infty \ \text{ if $m$ produced no field on $c$} .
  \end{aligned}
  \label{eq:blockscore}
\end{equation}

\paragraph{Errors are in meters, never in MOV pixels}
A MOV pixel is 0.6\,m at one extreme of the benchmark and 90\,m at the
other (Table~\ref{tab:legs}). Pooled in pixels, the same metric error
would therefore enter the aggregate with a $150\times$ different weight
depending on the cell it came from; in meters, every cell counts
equally.

\paragraph{Cross-cell aggregation is the median, never the mean}
Per-cell error is unbounded, so a single diverged cell can inflate a
mean (or RMS) arbitrarily far, leaving nothing meaningful to compare;
the median is unaffected. The failure tail is reported instead by each
cell's own p90, itself aggregated across cells by the same median.
This \emph{median per-cell p90} accompanies every reported median
in Section~\ref{sec:results}.

\paragraph{A cell with no field scores $+\infty$ and is still
aggregated}
Dropping unanswered cells would score each method only on the cells it
answered, rewarding failure: on one leg a method
answered 18 of 84 cells and, with the other 66 removed, ranked first.
Kept as $+\infty$, declines need no penalty constant: because every
reported aggregate is a
cross-cell median, a method's entry reads \textbf{FAIL} exactly when it
declines more than half of the cells it covers, in both the median and
the median per-cell p90.

\subsection{Baselines and fairness}
\label{sec:benchmark-baselines}

Seven training-free baselines span the classical families: SIFT+RANSAC
\cite{lowe2004sift,fischler1981ransac}, AROSICS \cite{scheffler2017arosics},
autoRIFT \cite{lei2021autorift}, GeFolki \cite{brigot2016gefolki}, RIFT
\cite{li2020rift}, RIFT2 \cite{li2023rift2}, and SRIF \cite{li2023srif}.
All run at official or published defaults on identical inputs,
and are scored by the identical evaluator. \method{} likewise runs one
configuration (Table~\ref{tab:selfcal}) on all 584 cells. RIFT and RIFT2
are faithful ports of the official MATLAB code, retaining the authors' FSC
outlier rejection; SRIF is reimplemented from its paper because no source
is published. Our port proved non-functional and is therefore
\emph{excluded from all rankings}, reported only for completeness and
flagged throughout (\excluded; \supp{II}).

Two pretrained matchers, SuperPoint+LightGlue
\cite{detone2018superpoint,lindenberger2023lightglue} and LoFTR
\cite{sun2021loftr}, were run zero-shot over the full matrix under a
documented adaptation layer (\supp{VI}). They are reported on E1 only
(Table~\ref{tab:e1-structural}) and are not ranked against the
training-free methods: they are not drop-in substitutes (a GPU, per-image
normalization, tiling), and their correspondences are densified through
our own global homography path, so ranking them as complete methods
would attribute our adaptation layer's limits to them.

\subsection{The ratio to the best}
\label{sec:benchmark-stats}

Beyond the block scores of Equation~\eqref{eq:blockscore}, the paper's
central claim of a bounded worst case needs a statistic that compares
a method to what was achievable on each fixture rather than to an
absolute standard. We use the ratio to the best method on the same block,
\begin{equation}
  R_m(B) \;=\; E_m(B) \;\big/\; \min_{m'} E_{m'}(B) ,
  \label{eq:ratio}
\end{equation}
and its per-cell analogue $r_m(c) = e_m(c) / \min_{m'} e_{m'}(c)$. Both
need no threshold, so no method is penalized on a block where every method
struggles, and the unknown b5 floor cancels. Their limitation is that
neither can tell ``all methods succeeded'' from ``all methods failed''
without the absolute table alongside, which is why both are always
reported next to meters. The per-cell form is used on \textbf{E2 only}:
seven E1 cells have a best of exactly zero, since the zero-displacement
constructions admit a method returning an exactly identity field, and
the denominator degenerates there.

% !TEX root = ../../SCDF.tex
\section{Results}
\label{sec:results}

All numbers come from the frozen 5840-run matrix, scored once under the rule
of Section~\ref{sec:benchmark-metrics}.
We first establish what the two experiments respectively measure
(Section~\ref{sec:results-scope}), then show that every baseline is a
specialist with a cliff (Section~\ref{sec:results-specialists}) and that
\method{} has no cliff and, as the same design predicts, no peak either
(Section~\ref{sec:results-bounded}).

\subsection{What each experiment measures}
\label{sec:results-scope}

% Table: E1, the structural test (all four blocks, all ten methods).
\begin{table*}[t]
\centering\small
\setlength{\tabcolsep}{4pt}
\caption{E1, the Structural Test: Block Median\,/\,p90 EPE (m)}
\label{tab:e1-structural}
\begin{threeparttable}
\begin{tabular}{@{}lrrrrrr@{}}
\toprule
 & \multicolumn{4}{c}{Block median\,/\,p90 EPE (m)}
 & Overall & Cells \\
\cmidrule(lr){2-5}
Method & A1 radiom. & A2 homog. & A3 resol. & A4 seam
       & median\,/\,p90 (m) & scored \\
\midrule
\method{} (ours)
  & 0.80\,/\,1.85 & 0.50\,/\,1.03 & 0.96\,/\,1.47 & 0.57\,/\,1.02
  & \textbf{0.63}\,/\,1.27 & 100/100 \\
GeFolki
  & 2.21\,/\,3.97 & 0.56\,/\,1424 & 1.17\,/\,2.04 & \textbf{0.26}\,/\,0.64
  & 1.02\,/\,3.73 & 100/100 \\
AROSICS
  & \textbf{0.39}\,/\,0.72 & 874.5\,/\,1561 & \textbf{0.28}\,/\,0.48 & 528.6\,/\,741.3
  & 1.29\,/\,4.07 & 99/100 \\
SIFT+RANSAC
  & 0.67\,/\,1.59 & \textbf{0.06}\,/\,0.50 & 7.11\,/\,7.51 & 446.2\,/\,907.1
  & 2.53\,/\,7.03 & 100/100 \\
autoRIFT
  & 3.06\,/\,7.15 & 579.3\,/\,1276 & 1.83\,/\,2.65 & FAIL
  & 4.02\,/\,28.0 & 78/100 \\
RIFT
  & 13.4\,/\,31.6 & 285.9\,/\,983.9 & 2.00\,/\,2.62 & 467.7\,/\,945.5
  & 49.6\,/\,343 & 100/100 \\
RIFT2
  & 7.59\,/\,9.31 & 729.1\,/\,2045 & 2.00\,/\,2.44 & 484.7\,/\,975.3
  & 19.8\,/\,29.9 & 100/100 \\
SRIF\excluded
  & 39933\,/\,61501 & 773.1\,/\,1980 & 3.64\,/\,4.66 & 482.5\,/\,976.5
  & 489\,/\,979 & 100/100 \\
\midrule
\multicolumn{7}{@{}l@{}}{\emph{Zero-shot pretrained matchers
  (Section~\ref{sec:benchmark-baselines}); not ranked against the above}} \\
LoFTR
  & 0.49\,/\,0.87 & 0.40\,/\,0.56 & 3.50\,/\,4.43 & 455.6\,/\,912.3
  & 1.54\,/\,3.42 & 100/100 \\
SuperPoint+LightGlue
  & 3.03\,/\,7.66 & 1.00\,/\,2.24 & 1.22\,/\,1.83 & 467.6\,/\,919.8
  & 1.84\,/\,5.03 & 100/100 \\
\bottomrule
\end{tabular}
\begin{tablenotes}[flushleft]\scriptsize\setlength\labelsep{0pt}
\item[] Each entry: block median\,/\,p90 EPE. Median $=$ the block score
  of Equation~\ref{eq:blockscore}; p90 $=$ the \emph{median per-cell p90},
  the cross-cell median of each cell's own per-pixel p90
  (Section~\ref{sec:benchmark-metrics}). 25 sites per block, exact ground
  truth, leg b1 throughout; 1\px{} $=$ 10\,m on this leg (30\,m under A3).
  \textbf{Bold} $=$ best block median in column. FAIL $=$ the block median
  falls on declined cells; \excluded{}~excluded from ranking.
\end{tablenotes}
\end{threeparttable}
\end{table*}

Tables~\ref{tab:e1-structural} and \ref{tab:e2-blocks} answer different
questions; E1 and E2 are not two difficulty levels of one test. In
Table~\ref{tab:e1-structural}, wherever a construction does
not violate a method's assumptions the block median lies between 0.06 and
7.1\,m, which the leg's GSD (10\,m; 30\,m under the $3\times$-coarser A3)
puts at 0.006--0.31\px{}, sub-pixel for every method. With exact
ground truth and no real content change, precision is not scarce,
so nothing in E1 ranks methods by it. E1 is a structural test. It
separates which assumption violation is fatal to which method, and its
large entries indicate a model class with no solution rather than a poor
score: AROSICS at 874\,m on A2, SIFT+RANSAC at 446\,m on A4.

% Table: E2, the realism test (by leg and by construction). The two panels
% marginalise the 20 blocks mapped in Figure~\ref{fig:heatmap}.
\begin{table*}[t]
\centering\footnotesize
\setlength{\tabcolsep}{3pt}
\caption{E2, the Realism Test: Block Median\,/\,p90 EPE (m) by Leg and by
Construction}
\label{tab:e2-blocks}
\begin{threeparttable}
\begin{tabular}{@{}lrrrrr@{\hspace{1.6em}}rrrr@{}}
\toprule
 & \multicolumn{5}{c@{\hspace{1.6em}}}{by leg (median\,/\,p90, m)}
 & \multicolumn{4}{c}{by construction (median\,/\,p90, m)} \\
\cmidrule(lr){2-6}\cmidrule(lr){7-10}
Method & b1 & b2 & b3 & b4 & b5 & A1 & A2 & A3 & A4 \\
       & 0.5\,yr & 1\,yr & 5\,yr & $\times$LS & $\times$NAIP
       & radiom. & homog. & resol. & seam \\
\midrule
\method{} (ours)
  & \textbf{3.26}\,/\,6.58 & \textbf{2.44}\,/\,4.54 & \textbf{5.10}\,/\,7.60
  & \textbf{11.5}\,/\,19.1 & \textbf{4.67}\,/\,7.77
  & 4.37\,/\,7.85 & \textbf{4.07}\,/\,7.62 & \textbf{3.89}\,/\,6.62
  & \textbf{4.36}\,/\,9.25 \\
GeFolki
  & 6.57\,/\,17.1 & 4.65\,/\,11.0 & 7.38\,/\,16.2 & 13.7\,/\,55.9
  & 5.43\,/\,9.34
  & 5.67\,/\,11.1 & 13.0\,/\,1366 & 7.07\,/\,15.5 & 4.73\,/\,9.72 \\
AROSICS
  & 19.1\,/\,69.6 & 15.6\,/\,44.5 & 15.8\,/\,69.4 & 80.3\,/\,378
  & 8.13\,/\,15.6
  & \textbf{4.01}\,/\,4.92 & 823.5\,/\,1511 & 4.48\,/\,6.24
  & 482.1\,/\,534.3 \\
SIFT+RANSAC
  & 9.23\,/\,19.0 & 6.87\,/\,10.1 & 11.5\,/\,16.8 & 39.2\,/\,74.8
  & 2032\,/\,3292
  & 8.53\,/\,17.1 & 6.69\,/\,9.39 & 11.2\,/\,19.6 & 458.4\,/\,918.5 \\
autoRIFT
  & 491\,/\,1105 & 491\,/\,1103 & 241\,/\,275 & 14.1\,/\,30.2 & FAIL
  & 7.48\,/\,12.5 & 587.9\,/\,1284 & 4.72\,/\,7.94 & FAIL \\
RIFT
  & 129\,/\,570 & 111\,/\,714 & 167\,/\,606 & 119\,/\,375 & 2400\,/\,4583
  & 48.4\,/\,92.3 & 287.1\,/\,1066 & 6.97\,/\,11.4 & 470.6\,/\,951.0 \\
RIFT2
  & 287\,/\,957 & 123\,/\,972 & 320\,/\,972 & 271\,/\,783 & 2443\,/\,3801
  & 40.3\,/\,54.2 & 795.2\,/\,2013 & 7.76\,/\,11.9 & 484.5\,/\,978.5 \\
SRIF\excluded
  & 32504\,/\,53441 & 689\,/\,1524 & 20016\,/\,29530 & 592\,/\,1230
  & 2660\,/\,4247
  & 41097\,/\,64243 & 1901\,/\,3100 & 16.7\,/\,21.2 & 501.7\,/\,985.1 \\
\midrule
Native floor $r_0$ (lit.)
  & $\approx$3--5 & $\approx$3--5 & $\approx$3--5 & $<$6 & --- & & & & \\
\bottomrule
\end{tabular}
\begin{tablenotes}[flushleft]\scriptsize\setlength\labelsep{0pt}
\item[] Each entry: block median\,/\,p90 EPE; p90 $=$ the \emph{median
  per-cell p90}, the cross-cell median of each cell's own per-pixel p90
  (Section~\ref{sec:benchmark-metrics}). 484 real-pair cells; the 20
  individual blocks marginalised here are in Figure~\ref{fig:heatmap}.
  \textbf{Bold} $=$ best block median in column; FAIL $=$
  the block median falls on declined cells; \excluded{}~excluded from
  ranking. Native floor $r_0$ is each pair's pre-existing misregistration
  as documented in literature (Section~\ref{sec:benchmark-construction}):
  same-relative-orbit Sentinel-2 pairs average 0.2--0.4\px{}
  ($\approx$3--5\,m) \cite{skakun2017coregistration,yan2018sentinel};
  Landsat Collection-2 is registered to the Sentinel-2 GRI to below 6\,m
  CE90 \cite{rengarajan2024coregistration}; b5's floor is undocumented
  (---), so its column ranks fairly but its absolute level is not an
  accuracy claim.
\end{tablenotes}
\end{threeparttable}
\end{table*}

E2 applies the same four constructions to real pairs
(Table~\ref{tab:e2-blocks}). Genuine content change now enters the
matching, and each pair's native misregistration $r_0$ affects every score
identically (Section~\ref{sec:benchmark-construction}). Here precision
is scarce, and the ranking is meaningful. \method{} spans
2.4--11.5\,m across the five legs and 3.9--4.4\,m across the four
constructions, and GeFolki, the steadiest baseline, spans 4.7--13.7\,m by
leg. Every other baseline loses at least one leg or construction; these
are the cliffs Section~\ref{sec:results-specialists} examines.

\subsection{Every baseline is a specialist, and every specialist has a cliff}
\label{sec:results-specialists}

% Figure: per-pixel EPE ECDFs on one cell.
\begin{figure}[tb]
  \centering
  \includegraphics[width=\linewidth]{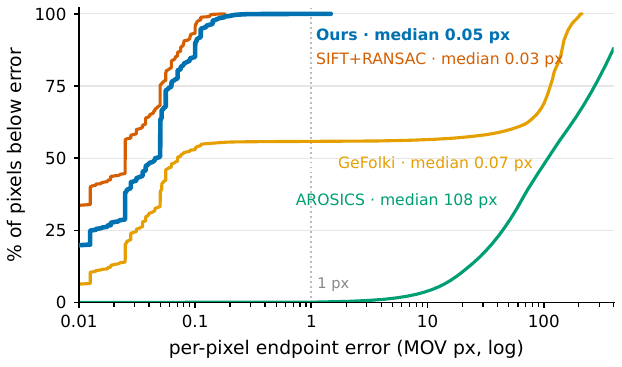}
  \caption{Per-pixel EPE ECDFs of four methods on one cell (E1/b1 $\cdot$
  A2 $\cdot$ \texttt{africa\_south\_1}). The axis is in MOV pixels because
  the figure stays inside a single cell; 1\px{} $=$ 10\,m on this leg.}
  \label{fig:ecdf}
\end{figure}

\begin{figure*}[t]
  \centering
  \includegraphics[width=\linewidth]{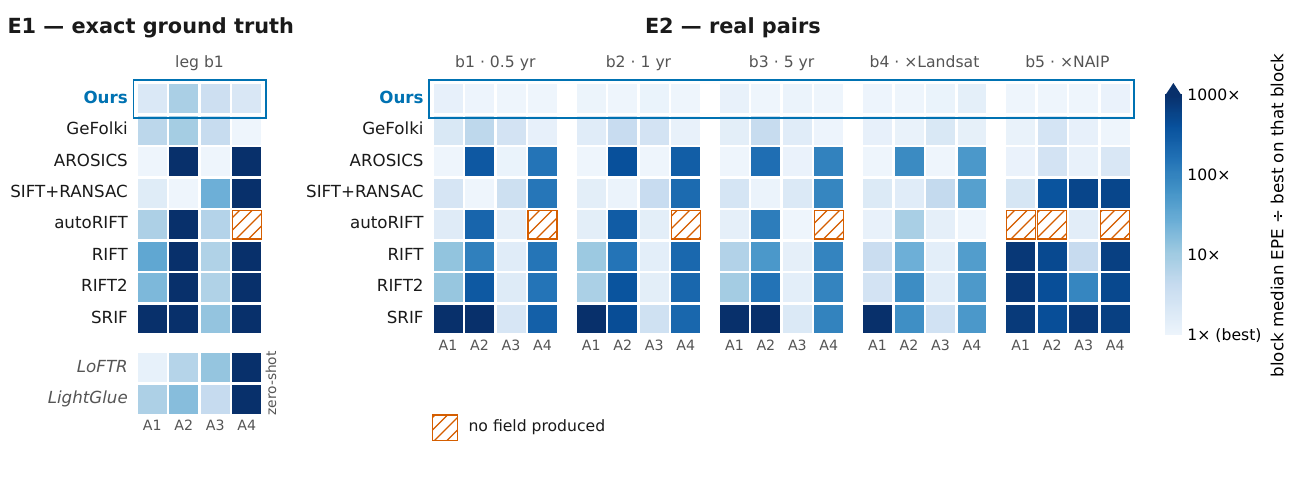}
  \caption{All 24 blocks, each method colored by its ratio to the best
  result on that block ($R_m(B)$, Equation~\eqref{eq:ratio}); absolute
  meters are in Tables~\ref{tab:e1-structural} and \ref{tab:e2-blocks}.
  Hatched: no field produced (a different outcome from a large error).}
  \label{fig:heatmap}
\end{figure*}

Figure~\ref{fig:heatmap} colors all 24 blocks by how far each method is from
the best result obtained on that block. Read by row, every
baseline row contains at least one dark block, and each dark block
corresponds to a mechanism the benchmark was built to expose.

% Figure: the cliff map --- every block, relative to that block's best.

\begin{itemize}
\item \textbf{SIFT+RANSAC} is the most accurate method in the paper on E1/A2,
at 0.062\,m (0.006 of a pixel, the bold A2 entry of
Table~\ref{tab:e1-structural}), because A2 is a homography and a
homography is exactly its model. The same model is 446\,m wrong on E1/A4
and 7.1\,m on E1/A3 (same table), and 458\,m on E2/A4
(Table~\ref{tab:e2-blocks}). The method is not weak; its accuracy depends on whether the construction
satisfies its single assumption.
\item \textbf{AROSICS} wins two of E1's four blocks (bold in
Table~\ref{tab:e1-structural}) and six of E2's twenty, more block wins
than any method except \method{}. It is also 823\,m off on E2/A2 and 482\,m on
E2/A4 (Table~\ref{tab:e2-blocks}), and it is $\ge 10\times$ from the best
in 8 of 20 E2 blocks; the E2 win and $\ge 10\times$ counts are the
placement columns of Table~\ref{tab:headline}. Per-window translation
estimation is accurate where the deformation is a translation and has no
representation for anything else.
\item \textbf{GeFolki} is the most consistent baseline (worst block
5.3$\times$, never $\ge 10\times$ off; Table~\ref{tab:headline}) and
wins E1/A4. Its weakness is a partial-failure
geometry: on E1/A2 its block median is 0.559\,m while its per-cell p90
reaches 1424\,m (the 0.56\,/\,1424 entry of Table~\ref{tab:e1-structural}),
i.e.\ it registers the low-displacement center of the warp accurately and
diverges at the corners. This is the behavior Figure~\ref{fig:ecdf} was
drawn from.
\item \textbf{RIFT and RIFT2} are 48\,m and 40\,m on the E2 radiometric
axis (the A1 column of Table~\ref{tab:e2-blocks}), an order of
magnitude behind methods with no radiometric design goal: their
phase-congruency descriptors target cross-modal contrast reversal, a
different problem from the gain/bias/blur perturbation of a same-optical
pair.
\end{itemize}

\textbf{The A4 seam, and autoRIFT's search window}:
A4 shifts the two halves of the scene in opposite directions, so no single
global transform has a solution, and the numbers are categorical rather
than graded. On E2's three same-sensor legs the global-model methods are at
430--615\,m while the two dense methods resolve the same cells to
2.3--5.2\,m (the per-leg blocks behind Table~\ref{tab:e2-blocks}'s A4
marginals, mapped in Figure~\ref{fig:heatmap}). No parameter can close
that factor of roughly 100, because the failure is mathematical.

autoRIFT is the informative exception: it is a dense method, but it searches
within a fixed 25\px{} window (the package default), and its A4 behavior
follows that envelope rather than the difficulty of the cell. It declines
94 of 96 A4 cells on the $8192^2$ legs, where the seam offset is
$\pm 49$\px{} (these declines are the A4 \textbf{FAIL} entries of
Tables~\ref{tab:e1-structural} and \ref{tab:e2-blocks}), and completes
\textbf{25 of 25} on E2/b4-A4, where the Landsat mover's $2731^2$ grid puts
the same offset at $\pm 16$\px, inside the window. On that one block it is
the best method in the benchmark at 8.84\,m; \supp{II} confirms the
envelope reading with a grid-size cross-check. A design envelope is
legitimate, but it is a commitment: correct use means sizing the search
window per dataset, from prior knowledge of how large the displacement can
be, so autoRIFT's reliability depends on the kind of hand-set constant that
\method{} is built not to carry.

\subsection{A bounded worst case, and no peak}
\label{sec:results-bounded}

% Table: bounded worst case.
\begin{table*}[t]
\centering\small
\setlength{\tabcolsep}{5pt}
\caption{Placement Across Blocks: The Bounded Worst Case}
\label{tab:headline}
\begin{tabular}{@{}lrrrrrrr@{}}
\toprule
 & Cells & \multicolumn{2}{c}{Overall (m)}
 & \multicolumn{4}{c}{Placement across blocks} \\
\cmidrule(lr){3-4}\cmidrule(lr){5-8}
Method & scored & median & p90 & best in & top-2 in & \textbf{worst ratio}
       & $\ge 10\times$ in \\
\midrule
\multicolumn{8}{@{}l@{}}{\textbf{E1} --- 100 cells, 4 blocks, exact ground truth} \\
\method{} (ours)     & 100/100 & \textbf{0.63} & \textbf{1.27} & 0 & 2 & \textbf{8.0}$\times$ & 0 \\
GeFolki              & 100/100 & 1.02 & 3.73 & 1 & 1 & 8.9$\times$ & 0 \\
AROSICS              &  99/100 & 1.29 & 4.07 & 2 & 2 & 13993$\times$ & 2 \\
SIFT+RANSAC          & 100/100 & 2.53 & 7.03 & 1 & 1 & 1697$\times$ & 2 \\
autoRIFT             &  78/100 & 4.02 & 28.0 & 0 & 0 & FAIL & 2 \\
RIFT2                & 100/100 & 19.8 & 29.9 & 0 & 0 & 11665$\times$ & 3 \\
RIFT                 & 100/100 & 49.6 & 343  & 0 & 0 & 4575$\times$ & 3 \\
SRIF\excluded        & 100/100 & 489  & 979  & 0 & 0 & 101302$\times$ & 4 \\
\addlinespace[1pt]
\multicolumn{8}{@{}l@{}}{\quad\emph{zero-shot, not ranked
  (Section~\ref{sec:benchmark-baselines})}} \\
\quad LoFTR          & 100/100 & 1.54 & 3.42 & 0 & 2 & 1733$\times$ & 2 \\
\quad LightGlue      & 100/100 & 1.84 & 5.03 & 0 & 0 & 1779$\times$ & 2 \\
\midrule
\multicolumn{8}{@{}l@{}}{\textbf{E2} --- 484 cells, 20 blocks, real pairs} \\
\method{} (ours)     & 484/484 & \textbf{4.17} & \textbf{7.77} & \textbf{10} & \textbf{19} & \textbf{1.39}$\times$ & \textbf{0} \\
GeFolki              & 484/484 & 6.83 & 17.8 & 1 & 7 & 5.3$\times$ & 0 \\
AROSICS              & 455/484 & 12.1 & 21.1 & 6 & 9 & 408$\times$ & 8 \\
SIFT+RANSAC          & 477/484 & 15.5 & 27.2 & 1 & 3 & 554$\times$ & 7 \\
autoRIFT             & 334/484 & 40.0 & 300  & 2 & 2 & FAIL & 9 \\
RIFT                 & 484/484 & 214  & 793  & 0 & 0 & 769$\times$ & 13 \\
RIFT2                & 484/484 & 480  & 975  & 0 & 0 & 783$\times$ & 13 \\
SRIF\excluded        & 484/484 & 2275 & 3658 & 0 & 0 & 16699$\times$ & 16 \\
\bottomrule
\addlinespace[2pt]
\multicolumn{8}{@{}l@{}}{\footnotesize Overall p90 $=$ the \emph{median
  per-cell p90}: the cross-cell median of each cell's own per-pixel p90
  (Section~\ref{sec:benchmark-metrics}).} \\
\multicolumn{8}{@{}l@{}}{\footnotesize \emph{Worst ratio} $= \max_B
  R_m(B)$, Equation~\eqref{eq:ratio}: the method's worst block relative to
  the best score achieved there.} \\
\multicolumn{8}{@{}l@{}}{\footnotesize FAIL $=$ a block median falls on
  declined cells; \excluded{}~excluded from ranking.} \\
\end{tabular}
\end{table*}

Table~\ref{tab:headline} is the paper's central result, and its two panels
have to be read together, because the design of
Section~\ref{sec:method-principle} predicted both.

On \textbf{E1} (the upper panel of Table~\ref{tab:headline}),
\method{} is best in none of the four blocks. It places
second, third, second and fourth (read down the block columns of
Table~\ref{tab:e1-structural}); AROSICS, SIFT+RANSAC and GeFolki each take a
block. Yet \method{} has the
lowest overall median (0.63\,m against GeFolki's 1.02\,m) and the lowest p90
(1.27\,m against 3.73\,m; the Overall columns). \method{} leads without
winning a block, as the only
method that is never far off: its worst block is 8.0$\times$ the best result
on that block (the \emph{worst ratio} column), which is E1/A2, where the
best result is SIFT's exact-model
0.062\,m and \method{}'s 0.500\,m is still one-twentieth of a pixel (the A2
column of Table~\ref{tab:e1-structural}). Two
methods have no block $\ge 10\times$ from the best; the other eight have two
to four (the $\ge 10\times$ column).

On \textbf{E2} (the lower panel), the same property becomes decisive,
because with 20 blocks
there are many more occasions for a catastrophic failure. \method{} is best in 10 of
20 blocks, top-two in \textbf{19 of 20}, and its worst block is
\textbf{1.39$\times$} the best available result. That block is E2/b4-A4,
where the method it trails is autoRIFT inside the one grid geometry that fits
its search window. For comparison, the best baseline's worst block is
5.3$\times$; AROSICS, which wins six blocks, is 408$\times$ off on another;
SIFT+RANSAC 554$\times$; RIFT and RIFT2 are $\ge 10\times$ from the best in 13
of 20 blocks each. \method{} produced a field on 584 of 584 cells; autoRIFT
declined 172, AROSICS 30, SIFT+RANSAC 7 (the cells-scored column, E1 and
E2 panels summed).

% Figure: per-cell ratio-to-best ECDF. E2 only — on E1 seven cells have a
% best of exactly zero and the ratio is undefined.
\begin{figure}[t]
  \centering
  \includegraphics[width=\linewidth]{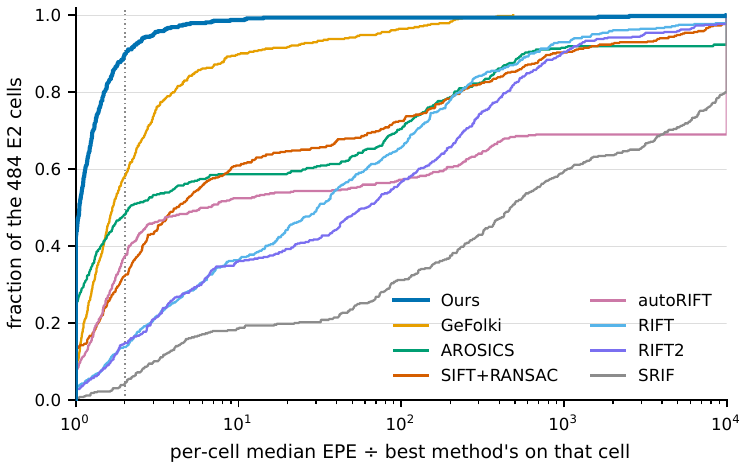}
  \caption{ECDF over the 484 E2 cells of each method's per-cell median EPE
  divided by the best obtained on that same cell ($r_m(c)$,
  Section~\ref{sec:benchmark-stats}). Left is better; dotted line:
  2$\times$; declined cells enter at the right edge, producing the plateaus
  below 1.0. E1 is omitted: seven of its cells have a best of exactly zero.}
  \label{fig:ratio-ecdf}
\end{figure}

Figure~\ref{fig:ratio-ecdf} shows the same property at cell resolution, where
it cannot be an artifact of block averaging. \method{} is within 2$\times$ of
the best available answer on 89.9\% of the 484 E2 cells and within 5$\times$
on 97.7\%; GeFolki reaches 58.3\% and 84.3\%, AROSICS 48.1\% and 56.6\%. Only
1.2\% of \method{}'s cells are more than 10$\times$ from the best, against
10.1\% for GeFolki and 41.3\% for AROSICS.

The claim this supports is narrower than ``most accurate'' and, we think, more
useful. When any competent method's assumptions hold, it is at least as
precise as \method{}: on E1 all of them are sub-pixel, and on E2 six
blocks go to AROSICS. No other method is free of a fixture
on which it is catastrophically wrong. For an unattended pipeline over
heterogeneous inputs, where the operator cannot know in advance which fixture
each pair resembles and cannot inspect every output, that property
determines whether the tool is usable.

The bound is a property of blocks; individual cells can still fail. On two
open-water sites, \texttt{east\_asia\_12} (b1/A1, b3/A1) and
\texttt{south\_asia\_9} (b3/A4), \method{} fails catastrophically, at
15.9--73.5\,km, and these three of the 484 E2 cells carry 92.9\% of its
total E2 error sum. Section~\ref{sec:discussion-limits} states the design
limit this failure family exposes, together with the fallback.

\subsection{What the components buy}
\label{sec:results-components}

% Table: factorial ablation of the measurement stack. Ratios are
% gmean(variant / full stack), so >1 means removing the switches costs
% accuracy; the full eight corners are in Supplement S-V.
\begin{table}[t]
\centering\small
\setlength{\tabcolsep}{3.5pt}
\caption{Factorial Ablation of the Measurement Stack on a Stratified
60-Cell Subset}
\label{tab:ablation}
\begin{tabular}{@{}lrrrr@{}}
\toprule
 & \multicolumn{2}{c}{Block score (m)} & \multicolumn{2}{c}{Ratio} \\
\cmidrule(lr){2-3}\cmidrule(l){4-5}
Configuration & median & p90 & median & p90 \\
\midrule
Full stack                   & 2.832 & 4.809 & 1.000 & 1.000 \\
$-$ sub-pixel re-measurement & 2.828 & 5.093 & 0.974 & 1.003 \\
$-$ lattice densification    & 2.899 & 5.484 & 1.014 & 1.093 \\
$-$ cross-validated fill     & 2.838 & 4.830 & 1.035 & 1.079 \\
$-$ all three                & 3.713 & 7.615 & \textbf{1.171} & \textbf{1.410} \\
\bottomrule
\end{tabular}
\par\vspace{3pt}
{\footnotesize\raggedright Ratio: geometric mean of per-cell
variant\,/\,full-stack error, on the per-cell median and p90; ${>}1$
means removing stages costs accuracy. The remaining corners and
interactions: \supp{V}.\par}
\end{table}

\paragraph{The measurement stack is a backstop} Taken one at a time, no
stage carries the precision: removing any single stage of the correlation
pass and fill costs at most 3.5\% on the typical cell
(Table~\ref{tab:ablation}). Removing all three at once costs 17.1\% on
the typical cell and 41.0\% on the per-cell tail, and the joint effect
exceeds the sum of the three individual effects by 14.7\%
(\supp{V}). The stages are mutually substitutable, each worth
little alone because the others compensate, so a
one-at-a-time ablation measures marginal value in a redundant set rather
than contribution. The stack pays where matches are imprecise: without it the
A1 radiometric cells are 1.53$\times$/1.83$\times$ worse (median/p90)
while the A2 homography cells move 1.004$\times$, and on every split the
tail moves more than the median.

No knockout changes whether a cell is registered: every variant,
the all-off corner included, produced a field on all 60 cells
(\supp{V}). The bounded worst case of
Section~\ref{sec:results-bounded} therefore does not come from any stage
the ablation can remove. It comes from the skeleton (the recall-first
policy, the leave-one-out filter, the null-calibrated response gate, the
confidence weighting), the components
Section~\ref{sec:method-principle} predicted, since they have no
constant to be wrong about. Although the skeleton cannot be switched off,
its response gate can be scored on its own: labeled against dense ground
truth on one cell's densification lattice, the per-window rule of
Section~\ref{sec:method-subpix} separates correct locks from false ones and matches
an oracle-tuned fixed threshold with no tuning (ROC and operating point
in \supp{V}).

\subsection{Computational cost}
\label{sec:results-runtime}

% Table: runtime on E2/b3, all methods pinned to one CPU core.
\begin{table}[t]
\centering\small
\caption{Single-Core Runtime on E2/b3}
\label{tab:runtime}
\begin{tabular}{@{}lrrr@{}}
\toprule
Method & $n/40$ & Median (s) & Min--max (s) \\
\midrule
SIFT+RANSAC   & 40 & 29   & 12--45 \\
autoRIFT      & 30 & 73   & 12--98 \\
AROSICS       & 40 & 150  & 57--1911 \\
SRIF\excluded & 40 & 316  & 139--538 \\
RIFT          & 40 & 330  & 48--394 \\
\method{} (ours) & 40 & 963 & 96--1540 \\
GeFolki       & 40 & 992  & 111--1260 \\
RIFT2         & 40 & 2174 & 173--2884 \\
\bottomrule
\end{tabular}
\par\vspace{3pt}
{\footnotesize\raggedright 10 sites $\times$ A1--A4; $8192^2$ inputs
($2731^2$ for A3); successful cells only; autoRIFT's ten missing cells
are its A4 failures. \excluded{}~Excluded from ranking.\par}
\end{table}

Estimating a dense displacement field is the expensive regime, and
\method{}'s cost is typical of it (Table~\ref{tab:runtime}): a median
of 963\,s per $8192^2$ scene (67\,Mpx), on par with GeFolki (992\,s) and
well under RIFT2 (2174\,s). Sparse global-model pipelines are far cheaper:
SIFT+RANSAC completes the same scenes in 29\,s, and autoRIFT needs 73\,s
because it searches only the fixed window of
Section~\ref{sec:results-specialists}. The cost is flat across the
full-size constructions and is dominated by the $K{=}8$ null correlations
and the dense correlation lattice. For comparability, every method in
Table~\ref{tab:runtime} is pinned to a single CPU core; in practice all of
them can spread across cores, and \method{} parallelizes trivially over
patches (${\sim}5\times$ wall-clock reduction with 8 workers,
bit-identical results), though no optimization was attempted for this
paper.

% !TEX root = ../../SCDF.tex
\section{Discussion}
\label{sec:discussion}

\subsection{Guidance for practice}

When a pair's regime is known in advance, the specialist that matches it
is the right tool, as our own tables show. A guaranteed same-sensor,
purely translational pair suits AROSICS, which is ${\sim}6\times$ faster
than \method{} and as precise (the A1/A3 columns of
Table~\ref{tab:e1-structural}); a pair known to be globally projective
suits SIFT+RANSAC (0.062\,m on E1/A2); a displacement known to be
small lets autoRIFT's window be sized to it, and inside that window it
is the best method we measured (E2/b4-A4). Prior knowledge can equally
be spent on tuning, since window sizes, grid spacings and model orders
are the constants it sets well. On a well-characterized dataset a tuned specialist
is a legitimate choice, and sometimes the better one.

\method{} is built for the complementary case, common under unattended
processing: heterogeneous or uninspected inputs whose regime nobody
certifies. There the specialists fail without warning, returning
fields hundreds of meters wrong with nothing in the output flagging the
failure (Section~\ref{sec:results-specialists}), and no constant can be
set correctly for a regime that is unknown. Every gate in \method{}
calibrates on the pair at hand, so there is no constant to set. What
transfers to unseen data is the bounded worst case of
Table~\ref{tab:headline}, not a win on any individual block. For evaluation
practice our recommendation is independent of our method and is stated in
Section~\ref{sec:benchmark-metrics}.

\subsection{Limitations of the method}
\label{sec:discussion-limits}

\paragraph{Open water} The catastrophic cells of
Section~\ref{sec:results-bounded} localize to open-water-dominated scenes.
Open water is a known difficulty for image matching in general and is not
particular to \method{}, because it carries no persistent structure to
match. The behavior at the cliff is particular to \method{}: on the
worst site the gates correctly detect the collapse (42 of 433 points
verify), but downweighting cannot recover a field in which no anchor is
reliable. A principled fallback, reverting toward the zero-displacement
seed when pair-level verification collapses, is future work; under the
$+\infty$ rule of Section~\ref{sec:benchmark-metrics} a declared
non-answer scores the same as a catastrophic field, so the fallback
would improve the product without improving our score.

\paragraph{Cost and scope} Runtime is unoptimized single-core
(${\sim}16$\,min per $8192^2$ scene serially; Section~\ref{sec:results-runtime}).
All development and evaluation used same-optical, same-band-family pairs.
Cross-modal registration is untested but not excluded: the matcher is
a plug point (Section~\ref{sec:method-cascade}), though we claim nothing
beyond the tested regime.

\subsection{Limitations of the benchmark, and disclosures}

All accuracy numbers in this paper rest on constructed ground truth. The
constructions are verified exact and are applied to real imagery, but each
leg's native floor (Section~\ref{sec:benchmark-construction}) bounds what
its numbers can claim: the b1--b3 floor ($\approx$3--5\,m) is taken from
literature rather than measured per pair; b4's is documented only as a
bound ($<6$\,m CE90); b5's is undocumented, so its columns rank methods
but support no absolute accuracy claim.

Three further disclosures follow, the first two detailed in
\supp{II}. autoRIFT's A4 failure is a design-envelope limit, confirmed
by a grid-size cross-check. SRIF is a non-functional reimplementation and
is excluded from every ranking. And single low-texture cells can swing
2--7$\times$ under sub-half-pixel perturbations, which is why no single
cell is ever cited as evidence.

% !TEX root = ../../SCDF.tex
\section{Conclusion}
\label{sec:conclusion}

We presented \method{}, a training-free dense displacement-field method for
co-registration of large optical satellite images in which every acceptance
decision self-calibrates from the image pair (per-window null-correlation
gates, pair-quantile error gates, studentized local-affine outlier tests, and
cross-validated interpolator selection), under one default configuration.
Across 584 constructed-ground-truth registration problems on real imagery,
precision separates the methods far less than the worst case does: with
exact ground truth every method is sub-pixel unless the fixture falls
outside its deformation model or its search envelope. On real
pairs \method{} produces a field on every cell and is never more than
1.39$\times$ from the best result on any block, while every baseline is somewhere 5.3--783$\times$ off or declines
cells. On non-global deformation the difference is structural, because
every global-model method (classical or zero-shot-learned with global
densification) is hundreds of meters wrong on cells that dense fields
resolve to a few meters. Standard mean/median-EPE reporting does not
expose any of this. For dense co-registration we propose scoring in
meters by order statistics that keep declined cells, read next to
distance-from-best on each fixture.

Future work: a principled fallback for pairs whose verification collapses
(open-water scenes); learned matchers behind the same self-calibrating gates;
manual-checkpoint evaluation on undistorted cross-sensor pairs as the
held-out anchor; true multimodal extension; and runtime engineering, which
was out of scope here. The matcher interface is a plug point,
and a dense learned matcher operating tile-wise could supply anchors where
RootSIFT cannot.

\paragraph{Data and code availability}
The source imagery, the site polygons with their WorldCover composition, and
the scripts that regenerate every constructed pair and its ground truth are
archived at \url{https://doi.org/10.5281/zenodo.21900160}. The implementation of
\method{} is at \url{https://github.com/TitorX/geocoreg}. The benchmark is regenerable end to
end from public Planetary Computer collections, since each construction is
deterministic given the archived scripts.

\bibliographystyle{IEEEtran}
\bibliography{references}

\end{document}

\title{Supplementary Material for:\\
Self-Calibrating Dense Displacement Fields for Reliable
Co-Registration of Large Optical Satellite Imagery}

% Names only; the main text carries the full author block.
\author{Shoukun~Sun, Zhe~Wang, Sanaz~Salati, Jiyin~Zhang, Hui~Wang,
        and~Xiaogang~Ma}

% Running heads.
\markboth{Supplementary Material}%
{Sun \MakeLowercase{\textit{et al.}}: Self-Calibrating Dense Displacement
Fields --- Supplementary Material}

\maketitle

\noindent This supplement collects implementation constants, baseline
configurations and audit disclosures, construction and data-pipeline details,
the component studies (gate ROC, factorial ablation), and the zero-shot deep-learning baseline details. Section,
table, and figure numbers here are prefixed with ``S''; references of the
form ``main-text Sect.~IV-B'' point to the main text.

\section{Method implementation details and constants}
\label{ssec:constants}

\subsection{Internal constants}

Table~\ref{tab:s-constants} lists the internal constants of the pipeline.
None of them is exposed as user configuration, and all are identical for
every one of the 584 benchmark cells.

\begin{table}[htbp]
\centering\small
\caption{Internal Constants of the \method{} Pipeline}
\label{tab:s-constants}
\begin{threeparttable}
\begin{tabular}{@{}ll>{\raggedright\arraybackslash}p{0.46\textwidth}@{}}
\toprule
Stage & Constant & Value \\
\midrule
Pyramid & initial patch size & overlap size \\
        & schedule & halve per level \\
        & minimum patch size & 256\px \\
        & center skip rule & skip if patch already holds $\geq 5$ matches \\
Matching & Lowe ratio threshold & 0.75 \\
         & local RANSAC & affine, patch-level \\
         & cycle consistency & forward/backward, max error 3\px \\
         & cross-resolution cap & 2048\px{} per side \\
         & resize inverse & center-correct, $p_{\mathrm{src}} = (p_{\mathrm{dst}} + 0.5)\,s - 0.5$, realized ratio $s$ \\
Per-level filters & global magnitude MAD gate & $k=3$ \\
                  & LOO local-affine neighbors & 12 \\
                  & robust reweight & one Tukey-IRLS pass ($c = 4.685$) \\
                  & leverage correction & residuals scaled by $1/\sqrt{1+g^2}$,
                    $g^2 = \sum_j a_j^2$ over the LOO prediction weights \\
                  & LOO threshold & $\mathrm{med} + 3.5 \cdot 1.4826\cdot\mathrm{MAD}$ of pair residuals \\
                  & NMS radius & 32\px{} (keep highest confidence) \\
Finest level & re-measurement window & 128 coarse px \\
        & densification window & 64 coarse px \\
        & lattice spacing & 48 coarse px \\
        & correlation upsampling & $32\times$ ($\approx 1/32$ coarse px) \\
        & null correlations per window & $K = 8$ \\
        & response gate & response $>$ max of the window's $K$ nulls \\
        & error gate & quantile of the pair's own error distribution
          ($q = 0.5$ re-measuring, $0.4$ densifying), clamped to
          $[0.6, 0.97]$ \\
        & gate failures & matched points downweighted by
          $\operatorname{clip}(\rho/\mathrm{med}\,\rho_{\mathrm{acc}}, 0.1, 1)$;
          lattice candidates never added \\
        & prior-deviation bound & 4 coarse px re-measuring, 3 densifying \\
Confidence $C$ & feature points & ratio-test confidence $\times$ reprojection confidence \\
        & correlation-measured points & measured lock quality: response $\rho$, clipped to $[0.05, 0.95]$ \\
        & consumers & MLS weights (priors and fill), NMS order, filter weights, gate-failure downweight \\
Fill & cross-validation & 5-fold over the pair's control points \\
        & candidate interpolators & 7 (MLS $k/\alpha$ family, IDW, TPS) \\
Warp & geolocation arrays & center-anchored, \texttt{PIXEL\_OFFSET}$=0.5$ \\
     & resampling & nearest-neighbor (original DN) \\
\bottomrule
\end{tabular}
\end{threeparttable}
\end{table}

\subsection{Leave-one-out filter: two refinements}
\label{ssec:loofilter}

The leave-one-out (LOO) local-affine filter of main-text Sect.~III-C predicts
each matched point's displacement from its $k = 12$ nearest matched
neighbors (self excluded), with the same distance- and confidence-weighted
affine model the MLS prior uses (main-text Eq.~(3)), and tests the
residual against main-text Eq.~(4). Two refinements keep that test honest.

\paragraph{Leverage studentization} The LOO prediction is linear in the
neighbor values. With the weights of main-text Eq.~(3) normalized to
$w_j$, the weighted centroid $\mathbf{p}^{*} = \sum_j w_j\,\mathbf{p}_j$,
centered offsets $\hat{\mathbf{p}}_j = \mathbf{p}_j - \mathbf{p}^{*}$, and
moment matrix $\mathbf{A} = \sum_j w_j\,\hat{\mathbf{p}}_j
\hat{\mathbf{p}}_j^{\!\top}$, the closed-form solution evaluated at
$\mathbf{p}_i$ is $\hat{\mathbf{d}}_{-i} = \sum_j a_j\,\mathbf{d}_j$ with
$a_j = w_j\bigl(1 + (\mathbf{p}_i - \mathbf{p}^{*})^{\!\top}\mathbf{A}^{-1}
\hat{\mathbf{p}}_j\bigr)$, so the prediction's noise scales with
$g = (\sum_j a_j^2)^{1/2}$. Interior points
have small $g$; a point at the hull edge or in a sparse area, whose fit
extrapolates, has $g$ several times larger, and a threshold calibrated on
the interior bulk would cut such valid high-leverage points. Residuals are
therefore scaled by $1/\sqrt{1 + g^2}$ before the threshold is set. A
neighborhood whose affine system is degenerate (collinear neighbors) falls
back to the weighted mean, with $g^2 = \sum_j w_j^2$.

\paragraph{One robust reweighting pass} An outlier corrupts the fits of
\emph{its neighbors}, inflating their residuals, and at a displacement
discontinuity the two sides corrupt each other. After a first pass, each
neighbor's weight is multiplied by the Tukey biweight of its own first-pass
residual, $(1 - u^2)^2$ for $|u| < 1$ and $0$ otherwise, with
$u = r_j / (c\,\hat\sigma_r)$, $c = 4.685$ and $\hat\sigma_r$ the robust
scale ($1.4826\,\mathrm{MAD}$) of the first-pass residuals; the fits are
then recomputed once. Only a point that remains inconsistent with a cleaned
neighborhood stays above threshold.

\subsection{Correlation-pass formalism}
\label{ssec:corrformalism}

Formal definitions for the correlation measurement of main-text
Sect.~III-D; window sizes, the upsampling factor, and $K$ are in
Table~\ref{tab:s-constants}. The two windows cover the same ground
extent, the finer one area-downsampled to the coarse grid, so every
quantity below is in coarse pixels. Both windows are mean-removed and
Hann-tapered, $\tilde{M} = (M - \bar{M}) \odot h$ and likewise
$\tilde{R}$; the taper suppresses the DFT's wrap-around seam. Every
candidate shift $\mathbf{s}$ is scored at once by the correlation surface
\begin{equation}
  \Gamma(\mathbf{s}) \;=\; \sum_{\mathbf{u}}
    \tilde{R}(\mathbf{u})\,\tilde{M}(\mathbf{u} + \mathbf{s})
  \;=\; \mathcal{F}^{-1}\bigl[\,\widehat{R} \odot
    \overline{\widehat{M}}\,\bigr](\mathbf{s}) ,
  \label{eq:s-corr}
\end{equation}
where the right-hand side --- the correlation theorem, with
$\widehat{\;\cdot\;}$ the 2-D DFT --- is how the surface is evaluated.
The residual is its peak,
\begin{equation}
  \mathbf{s}^{\star} \;=\;
  \operatorname*{arg\,max}_{\mathbf{s}}\, \Gamma(\mathbf{s}) ,
  \label{eq:s-residual}
\end{equation}
located to $1/32$\px{} by re-evaluating Eq.~\eqref{eq:s-corr} on an
upsampled grid in a small neighborhood of the integer peak --- a
matrix-multiply DFT over that neighborhood only \cite{guizar2008subpixel};
the windows themselves are never upsampled.

The two gate quantities are byproducts of the same transform products.
The response
\begin{equation}
  \rho \;=\; \max_{\mathbf{s}}\,
  \mathcal{F}^{-1}\!\left[
    \frac{\widehat{R} \odot \overline{\widehat{M}}}
         {\bigl|\widehat{R} \odot \overline{\widehat{M}}\bigr|}
  \right](\mathbf{s})
  \label{eq:s-response}
\end{equation}
is the peak of the phase-only correlation: every frequency votes on the
shift with unit weight, so unanimous agreement piles into a single spike
($\rho \to 1$) while ambiguous, self-similar content spreads it flat. The
error
\begin{equation}
  \epsilon \;=\; \Bigl(1 - \Gamma(\mathbf{s}^{\star})^{2} \big/
  \bigl(\lVert\tilde{R}\rVert^{2}\,\lVert\tilde{M}\rVert^{2}\bigr)
  \Bigr)^{1/2}
  \label{eq:s-error}
\end{equation}
is the normalized mismatch left after the best alignment: $\rho$ measures
whether the lock is unique, $\epsilon$ whether the windows look alike ---
two stretches of open water look alike but lock nowhere, while a sharp
cross-sensor lock is unique though radiometrically nothing alike, which
is why the main text gates the two axes separately. The three acceptance gates of main-text Sect.~III-D are, with
constants from Table~\ref{tab:s-constants}: the \emph{response gate},
\begin{equation}
  \rho \;>\; \max_{k = 1, \dots, K} \rho^{\mathrm{null}}_{k} ,
  \label{eq:s-nullgate}
\end{equation}
where $\rho^{\mathrm{null}}_{k}$ is the response of the same REF window
against the MOV window of another measurement in the batch, $K{=}8$
partners drawn at row-major index distance ${>}n/4$ so that their
content is unrelated (the gate is skipped when the batch holds fewer
than 16 windows); the \emph{error gate}, $\epsilon \le
\operatorname{clip}(Q_q(\epsilon), 0.6, 0.97)$ with $Q_q$ the
$q$-quantile over the batch ($q = 0.5$ re-measuring, $0.4$
densifying) --- an absolute bound would be meaningless, since good
cross-sensor locks sit near $\epsilon \approx 0.9$; and the
\emph{prior-deviation bound}, $|s^{\star}_{x}|, |s^{\star}_{y}| \le 4$
coarse px re-measuring and $3$ densifying. A measurement failing any
of the three is rejected outright, never clipped. Both data-driven
thresholds are computed only after the whole batch is measured. In the
gate-failure downweight of Table~\ref{tab:s-constants},
$\rho_{\mathrm{acc}}$ denotes the responses of the windows accepted in
the same pass, so a failed point's confidence is scaled by its response
relative to the accepted median.

\section{Baseline configurations and audit disclosures}
\label{ssec:baselines}

\subsection{Configurations}

All baselines run at official or published defaults, on identical inputs,
scored by the identical evaluator. Single-CPU execution is
enforced at each method's entry point (OpenCV/BLAS/OpenMP thread pools set
to 1; AROSICS \texttt{CPUs}$=$1; autoRIFT \texttt{MultiThread}$=$0), verified
at 99.7\% single-core occupancy during the runtime measurement.

\begin{itemize}
\item \textbf{SIFT+RANSAC} (OpenCV): 8000 features, Lowe ratio 0.75, RANSAC
reprojection threshold 5\px, single global homography evaluated densely on
MOV's grid.
\item \textbf{AROSICS}: default global + local tie-point grid, window/grid
sizes at the package's adaptive defaults.
\item \textbf{autoRIFT}: official minimal configuration --- chip sizes
32--64\px{} and chip-grid spacing 32\px{} (the class defaults), oversampling
ratio 64 (the one value the official script overrides), \emph{search limit
25\px} (the package default), sparse chip displacements interpolated to the
full MOV grid.
\item \textbf{RIFT / RIFT2}: faithful function-by-function ports of the
official MATLAB code (phase congruency, FAST-on-moment-map keypoints,
maximum-index-map descriptors; patch size 96, 5000 keypoints), retaining the
authors' FSC outlier rejection (affine for RIFT, similarity for RIFT2)
rather than substituting RANSAC.
\item \textbf{SRIF}: reimplemented from the paper (no source is published;
the authors provide only a Windows executable); paper Table-3 parameters.
See the disclosure below.
\item \textbf{GeFolki}: the official ONERA Python implementation, adopted
unmodified, at its defaults.
\end{itemize}

The four ported methods (RIFT, RIFT2, SRIF, GeFolki) share one input
preparation step that warps REF onto MOV's exact grid first.
All methods emit a dense field on MOV's full grid; no method
warps its own output --- rendering for image-space scoring is one shared
routine.

\subsection{Disclosure: SRIF port is non-functional}
\label{ssec:srif}

Our SRIF reimplementation produces invalid geometry even on near-identity
input: on an E1/b1 A1 cell (radiometric-only distortion, true offset
$\approx 0$) it fits a homography with a $\approx 99^{\circ}$ rotation,
scale 0.68, and a 7259\px{} translation from 3 inliers (cell median EPE
$\approx 42$\,km); its frozen per-leg block medians span
0.49--32.5\,km, with 32.5\,km on a same-GSD leg. This does not represent the published
SRIF algorithm --- it is a defect of our port (the paper omits at least one
parameter we had to guess). SRIF rows are therefore retained in tables for
completeness, flagged $\dagger$, and excluded from every ranking, claim, and
aggregate in the paper.

\subsection{Disclosure: autoRIFT's A4 failure is its design envelope}
\label{ssec:autorift}

autoRIFT fails most A4 cells because the A4 seam displacement
($\pm 0.006\cdot\max(H,W)$) exceeds its official fixed 25\px{} search
window. When the true displacement is outside the window, the NCC finds no
peak, the refinement collapses the search limit to zero, the chip grid
returns NaN, and the wrapper reports a (correct) failure --- the mechanism
the tool's own documentation describes. The grid-size cross-check confirms
the mechanism: on the $8192^2$ legs autoRIFT completes 3/25 (E1/b1) and
2/96 (E2) A4 cells, whereas on E2/b4 the crops are $2731^2$, the seam is
only $\pm 16$\px{} --- inside the window --- and it completes 25/25. autoRIFT was designed for small, smooth glacier-velocity
fields; A4 is outside its envelope at official defaults. We deliberately did
not retune \texttt{search\_limit}: that would break the published-defaults
protocol. (E2/b5 is a whole-leg pathology for autoRIFT --- 0/21 completions
on A1/A2/A4, 18/21 on A3 --- recorded as failures, not dropped cells.)

\section{Construction details and ground-truth verification}
\label{ssec:constructions}

\subsection{Formalism, equations, and parameters}

All four constructions share one formalism. Let $B$ denote the base raster
(the source scene in E1, the real mover in E2), a square of side $N$\px,
and let $\varphi$ be the affine pixel-to-map transform of the constructed
mover's grid. A geometric construction is a pixel-space displacement field
$\mathbf{u}$; the mover $M$ and its ground truth follow as
\begin{equation}
  M(\mathbf{p}) \;=\; B\bigl(\mathbf{p} - \mathbf{u}(\mathbf{p})\bigr) ,
  \qquad
  \mathbf{D}_{\mathrm{gt}}(\mathbf{p}) \;=\;
  \varphi\bigl(\mathbf{p} - \mathbf{u}(\mathbf{p})\bigr) -
  \varphi(\mathbf{p}) ,
  \label{eq:s-construction}
\end{equation}
with bilinear sampling of $B$, so the truth is a map-unit field on the
mover's own grid --- exactly the convention of main-text Eq.~(1). A
mover pixel is valid only if its full bilinear footprint is in bounds and
touches no nodata sample. Distortions act on raw sample values and
preserve dtype and nodata; the reference is never touched. All magnitudes
are fixed fractions of $N$, and all random draws use one fixed seed, so
every site receives the identical realization; parenthetical pixel values
below instantiate $N = 8192$.

\begin{itemize}
\item \textbf{A1 (radiometric + PSF; ground truth $\equiv 0$).}
$\mathbf{u} \equiv \mathbf{0}$. Each band $b$ is tone-remapped on its own
raw value scale $s_b$ (the band's valid-pixel 98th percentile), blurred,
and re-noised:
\begin{equation}
  M_b \;=\; G_{\sigma} \ast
  \Bigl[\, s_b\,\bigl( g_b\,(B_b / s_b)^{\gamma} + \beta_b \bigr) \Bigr]
  \;+\; \varepsilon_b ,
  \qquad
  \varepsilon_b \sim \mathcal{N}\bigl(0,\,(0.04\,s_b)^{2}\bigr) ,
  \label{eq:s-a1}
\end{equation}
with tone exponent $\gamma = 1.6$, per-band gain
$g_b \sim \mathcal{U}(0.80, 1.20)$ and bias
$\beta_b \sim \mathcal{U}(-0.06, 0.06)$, and $G_{\sigma}$ a $\sigma =
2$\px{} Gaussian emulating an MTF difference, computed as a normalized
convolution over valid samples so that nothing bleeds across nodata; the
result is clipped to the native dtype.
\item \textbf{A2 (global homography).}
$\mathbf{u}(\mathbf{p}) = \mathbf{p} - \mathbf{H}^{-1}\mathbf{p}$, i.e.\
$M(\mathbf{p}) = B(\mathbf{H}^{-1}\mathbf{p})$, where the homography
$\mathbf{H}$ (acting in homogeneous coordinates) maps the four scene
corners onto themselves plus i.i.d.\ $\mathcal{U}(-\tau, \tau)^{2}$
offsets, $\tau = 0.02\,N$ ($\approx 164$\px{} at $8192^2$); the same
inverse homography that warps the image generates the ground-truth field.
\item \textbf{A3 (downsampling).} A grid change rather than a
displacement: $M$ is the nodata-aware area average of $B$ onto a coarse
grid of side $N' = \operatorname{round}(N/3)$, and the geotransform uses
the exact realized factor $N/N'$ --- under which the zero-displacement
ground truth is rigorously exact (MOV pixel-center positions match the
INTER\_AREA content centroids to 0.0\,m). Under the nominal factor 3 the
zero truth would acquire a linear ramp reaching $\approx 0.33$ MOV px at
the far edge; an early version wrote exactly that, was fixed before the
benchmark freeze, and all frozen data use the exact factor.
\item \textbf{A4 (seam-dominated non-global deformation).} A mosaic seam
plus a weak elastic residual,
\begin{equation}
  \mathbf{u}(\mathbf{p}) \;=\;
  \delta\,\tanh\!\Bigl(\frac{x - N/2}{N/16}\Bigr)\,\hat{\mathbf{x}}
  \;+\; \mathbf{e}(\mathbf{p}) ,
  \qquad \delta = 0.006\,N ,
  \label{eq:s-a4}
\end{equation}
so the left and right halves shift in opposite directions by $\delta$
($\pm 49$\px{} at $8192^2$) across a smooth $\tanh$ transition band. The
elastic residual $\mathbf{e}$ is generated as per-axis white noise of
standard deviation 24\px{} on a $16\times$-downsampled grid and
Gaussian-smoothed with a kernel proportional to image size --- the
smoothing attenuates the \emph{effective} amplitude to $\approx 0.41$\px{}
at $8192^2$ ($\approx 1.19$\px{} at $2731^2$).
\end{itemize}

For every construction the stored ground truth was verified against the
constructed image: resampling the base with the ground-truth field
reconstructs the constructed mover with 0.0 error for A2 and A4 (seam
discontinuities land identically in image and field); A1's truth is zero by
definition on an untouched grid; A3's zero-truth is exact under the
realized factor. Nodata footprints propagate conservatively into the valid
masks.

\section{Site sampling and data pipeline}
\label{ssec:pipeline}

Sites are sampled globally by area-weighted random draws over continental
land boxes (cosine-latitude corrected; ocean points are rejected naturally
by the absence of WorldCover tiles). Each candidate's ESA WorldCover 2021
v200 composition \cite{zanaga2022worldcover} is probed over the tier's
representative crop footprint ($\approx 82$\,km window for the
Sentinel-2/Landsat tier, $\approx 5$\,km for the NAIP tier, decimated to
$\approx 100$\,m), and accepted iff (i) the largest class covers
$\le 90\%$ and (ii) at least two of the seven main classes exceed 2\%
(``anti-homogeneity''), with a set-level target of $\ge 10$ sites per class;
built-up never dominates an 82-km window and is under-covered (4--5 sites)
--- an intrinsic gap we disclose. Accepted sites keep a minimum center
separation of one footprint side (twice the probe half-window), so crops
never overlap. Per-site class
fractions are stored in the released site polygons.

All imagery is fetched from Microsoft Planetary Computer
\cite{microsoft2022planetary} STAC collections (Sentinel-2 L2A, Landsat
Collection 2 L2, NAIP), reading cloud-optimized GeoTIFFs remotely. Pair
rules: scene-level cloud ${<}10\%$; Sentinel-2 multi-temporal pairs must
share the MGRS tile \emph{and the relative orbit} (the hard geometric
constraint protecting the L1 floor; cross-orbit pairs are excluded
outright); candidates are sorted by cloud cover and walked until both crops
pass quality control (crop-level invalid fraction $\le 20\%$ from the SCL
layer for Sentinel-2 --- cloud shadow, medium/high-probability cloud,
cirrus, snow --- the QA band for Landsat, and all-zero nodata for NAIP).
Time windows: E2/b1 reference 2024-07 $\pm 60$\,d versus mover 2024-01
$\pm 75$\,d; E2/b2 2023-07 $\pm 90$\,d; E2/b3 2019-07 $\pm 150$\,d; E2/b4
Sentinel-2 and Landsat-8/9 each within $\pm 30$\,d of 2024-07, picked
independently, so realized pair gaps run 0--50\,d (median 16\,d); E2/b5
Sentinel-2 within $\pm 15$\,d of 2024-07, paired with each site's
single-vintage NAIP mosaic --- the year with maximal tile coverage over the
site within 2017--2024, realized as 2022 (11 sites) or 2023 (10), same-year
tiles only to avoid cross-vintage seams. The b5 pairs therefore carry
0.7--2.1\,yr (median 1.9\,yr) of landscape change on top of the modality
gap.

\section{Component studies: gate ROC and factorial ablation}
\label{ssec:ablation}

\subsection{Response-gate ROC}

\begin{figure}[htbp]
  \centering
  \includegraphics[width=0.8\textwidth]{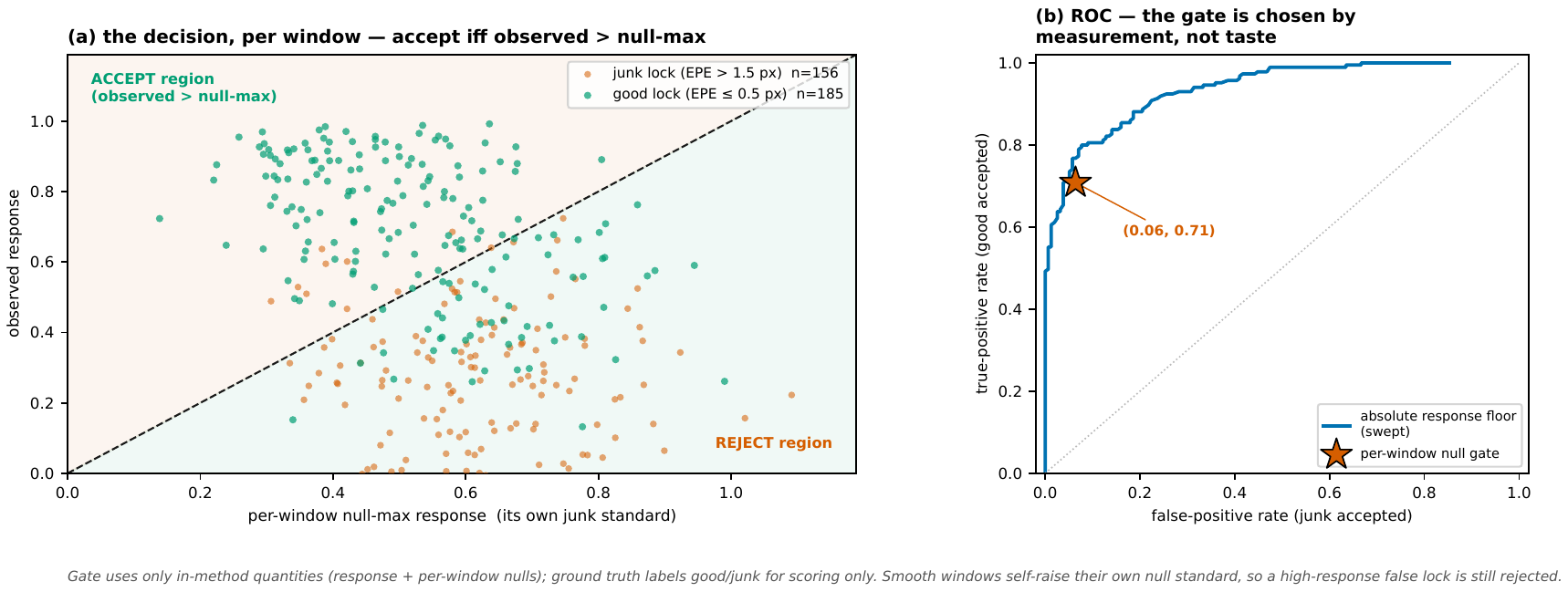}
  \caption{The response gate scored against dense ground truth, on the 342
  measurable densification-lattice windows of one cell
  (E2/b4$\cdot$A3$\cdot$\texttt{africa\_south\_8}). Ground truth labels
  windows good/junk for scoring only --- the gate sees none of it.
  (a)~The per-window accept rule. (b)~The gate's operating point against
  the frontier traced by sweeping an absolute response floor.}
  \label{fig:s-gate-eval}
\end{figure}

Figure~\ref{fig:s-gate-eval} scores the response gate against dense
ground truth on one of the probe cells on which it was selected. Each
densification-lattice window is labeled good (EPE $\le 0.5$ coarse px)
or junk (${>}1.5$) from the ground truth, for scoring only; the gate's
accept/reject decisions are scored as true- and false-positive rates and
compared with the ROC curve of an absolute response floor swept over its
full range.

The gate accepts 71\% of the good windows and 6\% of the junk ones
(TPR 0.71 / FPR 0.06), a point on the frontier of the swept floor: with
no tuning, it matches what a floor tuned on this cell's own ground truth
could achieve. Parity without tuning, not dominance, is the claim.
Panel~(a) shows why a single floor cannot do the same: the per-window
standard spans 0.14--1.13 across windows, so a smooth window is held to
a higher bar than a textured one.

\subsection{Factorial ablation of the measurement stack}
\label{ssec:factorial}

Design: a full $2^3$ factorial that switches each of the three stages
of the measurement stack on or off --- sub-pixel re-measurement,
lattice densification, and cross-validated fill, hereafter
\emph{re-measurement}, \emph{densification}, and \emph{fill} ---
giving eight variants from V0 (all on, the shipped configuration) to V7
(all off), each run on the same stratified 60-cell subset
(first sites per leg in sweep order $\times$ all four
constructions; 12 cells per leg group).
Cells are not independent --- each site recurs across legs and
constructions --- so the paired Wilcoxon $p$ values throughout this
section are read as descriptive strength, not literal error rates; the
main text reports no significance statistics.

Table~\ref{tab:s-ablation} gives the eight corners (V0--V3 and V7 are
main-text Table~VII, without the $p$ columns). Removing any one stage
costs at most 3.5\% on the median; removing all three costs 17.1\% on
the median and 41.0\% on p90, with every two-stage knockout in between.
No stage carries the precision alone.

\begin{table}[htbp]
\centering\small
\caption{The Eight Corners of the Factorial: Block Scores and Paired
Ratios to V0}
\label{tab:s-ablation}
\begin{threeparttable}
\begin{tabular}{@{}llrrrrrr@{}}
\toprule
 & & \multicolumn{2}{c}{Block score (m)}
 & \multicolumn{2}{c}{Paired on median} & \multicolumn{2}{c}{Paired on p90} \\
\cmidrule(lr){3-4}\cmidrule(lr){5-6}\cmidrule(lr){7-8}
Variant & Stages off & median & p90 & ratio & $p$ & ratio & $p$ \\
\midrule
V0 & --- (full stack)            & 2.832 & 4.809 & 1.000 & --- & 1.000 & --- \\
V1 & re-measurement               & 2.828 & 5.093 & 0.974 & 0.81 & 1.003 & 0.15 \\
V2 & densification                & 2.899 & 5.484 & 1.014 & 0.091 & 1.093 & $1.8\times10^{-5}$ \\
V3 & fill                         & 2.838 & 4.830 & 1.035 & 0.0034 & 1.079 & $1.5\times10^{-5}$ \\
V4 & re-measurement $+$ densification & 3.250 & 6.212 & 1.039 & 0.0014 & 1.171 & $1.4\times10^{-7}$ \\
V5 & densification $+$ fill       & 2.957 & 5.944 & 1.065 & $5.3\times10^{-4}$ & 1.218 & $2.2\times10^{-9}$ \\
V6 & re-measurement $+$ fill      & 3.120 & 5.938 & 1.061 & $1.7\times10^{-4}$ & 1.161 & $1.8\times10^{-8}$ \\
V7 & all three                   & 3.713 & 7.615 & \textbf{1.171} & $5.2\times10^{-7}$ & \textbf{1.410} & $5.4\times10^{-11}$ \\
\bottomrule
\end{tabular}
\begin{tablenotes}[flushleft]\footnotesize\setlength\labelsep{0pt}
\item[] Ratio: geometric mean over the 60 cells of per-cell variant/V0,
  on the per-cell median and on the per-cell p90; ${>}1$ means switching
  the stages off costs accuracy. $p$: paired Wilcoxon over the 60 cells.
  All variants registered all 60 cells at the same median coverage (0.947).
\end{tablenotes}
\end{threeparttable}
\end{table}

Table~\ref{tab:s-effects} decomposes the same runs into main effects
and pairwise interactions. All three main effects are positive on both
responses ($p \le 0.010$), including re-measurement, whose one-at-a-time
row V1 shows nothing (0.974, $p{=}0.81$) because densification and fill
cover for it: with both gone, removing it costs 10.0\%. All three
interactions are positive --- each stage is worth more when the others
are absent --- and the joint effect exceeds the sum of the three
single-stage effects by 14.7\% ($p{=}1.7\times10^{-9}$). The stages are
substitutes, not additive contributors; the redundancy the main text
asserts is measured here.

\begin{table}[htbp]
\centering\small
\caption{Main Effects and Interactions of the Three Stages}
\label{tab:s-effects}
\begin{threeparttable}
\begin{tabular}{@{}llrrrr@{}}
\toprule
 & & \multicolumn{2}{c}{Median} & \multicolumn{2}{c}{p90} \\
\cmidrule(lr){3-4}\cmidrule(lr){5-6}
Term & Pairs, off/on & ratio & $p$ & ratio & $p$ \\
\midrule
main: re-measurement & V1/V0, V4/V2, V6/V3, V7/V5 & 1.030 & 0.010 & 1.075 & $3.0\times10^{-5}$ \\
main: densification  & V2/V0, V4/V1, V5/V3, V7/V6 & 1.053 & $2.1\times10^{-5}$ & 1.150 & $1.8\times10^{-8}$ \\
main: fill           & V3/V0, V5/V2, V6/V1, V7/V4 & 1.075 & $2.3\times10^{-9}$ & 1.138 & $3.0\times10^{-11}$ \\
interaction: re-measurement $\times$ densification & V4/V2, V7/V5 vs.\ V1/V0, V6/V3 & 1.063 & $1.3\times10^{-5}$ & 1.072 & $8.3\times10^{-4}$ \\
interaction: re-measurement $\times$ fill          & V6/V3, V7/V5 vs.\ V1/V0, V4/V2 & 1.063 & $1.1\times10^{-6}$ & 1.077 & $1.1\times10^{-5}$ \\
interaction: densification $\times$ fill           & V5/V3, V7/V6 vs.\ V2/V0, V4/V1 & 1.025 & $1.2\times10^{-4}$ & 1.036 & $8.5\times10^{-5}$ \\
joint vs.\ combined single-stage effects (V1--V3)  & V7/V0 vs.\ V1/V0, V2/V0, V3/V0 & \textbf{1.147} & $1.7\times10^{-9}$ & \textbf{1.191} & $3.8\times10^{-7}$ \\
\bottomrule
\end{tabular}
\begin{tablenotes}[flushleft]\footnotesize\setlength\labelsep{0pt}
\item[] Per-cell contrasts on log error, geometric mean over the 60
  cells; $p$: paired Wilcoxon of the per-cell contrasts against zero.
  Pairs: a main effect averages the per-cell log ratios of its four
  off/on pairs; an interaction is the average over the pairs before
  ``vs.'' minus the average over those after; the joint row is V7/V0
  minus the sum of the three single-stage contrasts. Ratios are
  arithmetic transforms of Table~\ref{tab:s-ablation} (exact before
  rounding); $p$ values are not.
\end{tablenotes}
\end{threeparttable}
\end{table}

Table~\ref{tab:s-splits} splits the all-off cost V7/V0 by construction
and by leg (no clean-cell conditioning). The stack pays where matches
are imprecise: the radiometric axis A1 loses most without it
($1.53\times$ median, $1.83\times$ p90), since radiometric difference is
what makes feature localization imprecise and all three stages attack
exactly that, and the 1:3 cross-resolution leg b4 is the most damaged
leg on the median ($1.29\times$; $1.50\times$ on p90). A2 pays nothing
(1.004): homography matches are already precise, and re-measuring them
only adds noise --- V1 on A2 is 0.887 median, 0.874 p90, the one place
where switching re-measurement off helps. That A1$\uparrow$/A2$\downarrow$
cancellation is what the pooled V1 row averages to null. b5 moves least
(1.053) but is $r_0$-confounded (main-text Sect.~IV-B) and
uninformative here.

\begin{table}[htbp]
\centering\small
\caption{Where the Stack Matters: V7/V0 by Construction and by Leg}
\label{tab:s-splits}
\begin{threeparttable}
\begin{tabular}{@{}lrrrrcrrrrr@{}}
\toprule
 & A1 & A2 & A3 & A4 & & b1 & b2 & b3 & b4 & b5 \\
\midrule
V7/V0 on median & 1.527 & 1.004 & 1.106 & 1.110 & & 1.171 & 1.223 & 1.137 & 1.292 & 1.053 \\
V7/V0 on p90    & 1.831 & 1.108 & 1.499 & 1.298 & & 1.449 & 1.502 & 1.326 & 1.501 & 1.263 \\
\bottomrule
\end{tabular}
\begin{tablenotes}[flushleft]\footnotesize\setlength\labelsep{0pt}
\item[] Geometric mean of per-cell V7/V0 over the split's cells, on the
  per-cell median and p90.
\end{tablenotes}
\end{threeparttable}
\end{table}

Reliability is overwhelmingly --- not entirely --- unmoved. Under the
0.90 sub-pixel-fraction diagnostic (a cell counts as clean when at least
90\% of its pixels have EPE under one MOV pixel; a supplement-only
classification, since the main text defines no success threshold), the
clean-rate is 0.717 for every variant except the two deepest knockouts
(V4: 0.683, V7: 0.667, i.e.\ 2 and 3 cells): no single-stage knockout flips any
cell. The worst single-cell
degradation grows with knockout depth --- $1.57\times$ (V1),
$2.38\times$ (V4), $3.49\times$ (V7) --- and V0 beats V7 on 46 of 60
cells. The stages are precision components, not reliability components.

\section{Zero-shot deep-learning baselines}
\label{ssec:dl}

SuperPoint$+$LightGlue (official repository weights
\texttt{superpoint\_v1}, \texttt{superpoint\_lightglue\_v0-1}) and LoFTR
(outdoor weights via kornia 0.8.3 \cite{riba2020kornia}) run strictly
zero-shot, with published pretrained weights, default inference settings,
and no fine-tuning or domain adaptation.

Both models take single-channel 8-bit input and are memory-bound far
below the $8192^2$ scene size, while the benchmark rasters are multi-band
UInt16. The shared adaptation layer therefore (i)~collapses the bands to
grayscale by an unweighted mean, (ii)~rescales each image to 8 bits by a
2--98th-percentile stretch computed over its valid samples, with nodata
excluded from the statistics and written as black, and (iii)~cuts REF and
MOV into co-located tiles. When the two rasters already share a pixel
grid (like A1/2/4) the tiles are cut
directly; otherwise (like A3) REF is first
resampled onto MOV's grid by bilinear interpolation, after which
corresponding content falls in the same tile up to the pair's actual
misregistration, which the tile overlap absorbs. LightGlue uses $2048^2$
tiles with 256\,px overlap (25 tiles per $8192^2$ pair); LoFTR uses
$1024^2$ tiles with 128\,px overlap (81 tiles), because its dense coarse
attention is memory-bound in tile area.
Tile-local matches are aggregated to full-image coordinates and
densified by the same global-homography estimation as SIFT$+$RANSAC.

\bibliographystyle{IEEEtran}
\bibliography{references}